\RequirePackage[OT1]{fontenc}
\documentclass[a4paper,10pt,conference]{ieeeconf}
\usepackage{graphicx,amsmath,amssymb,booktabs,multirow,xcolor}
\usepackage[hidelinks]{hyperref}
\IEEEoverridecommandlockouts
\title{\LARGE\bf FatigueFormer: A Physiology-Informed Model\\for Upper-Limb Muscle Fatigue Assessment}
\author{Tong Zhang, Hong Guo\thanks{Hong Guo is the corresponding author.}, Shuangzhou Yan,
Dongkai Weng, Jian Wang, Hongxin Zhang\\
Zhejiang University, Hangzhou, China\\
{\small Tong Zhang: \texttt{12521180@zju.edu.cn}}}
\hypersetup{
  pdftitle={FatigueFormer: A Physiology-Informed Model for Upper-Limb Muscle Fatigue Assessment},
  pdfauthor={Tong Zhang, Hong Guo, Shuangzhou Yan, Dongkai Weng, Jian Wang, Hongxin Zhang}
}
\begin{document}
\pagestyle{plain}
\maketitle
\begin{abstract}
Surface electromyographic (sEMG) signals provide information about muscle activity and function during movement. Fatigue alters multiple properties of these signals, with patterns that vary across individuals and contraction intensities. However, physiological insights into fatigue remain underused in sEMG modeling, particularly for cross-subject recognition. We introduce FatigueFormer to recognize fatigue states defined by ratings of perceived exertion (RPE). The model incorporates physiologically motivated trend grouping into sEMG encoding by organizing descriptors into increasing- and decreasing-trend streams and combining inter-window context with within-window feature interactions. To assess fatigue-state discriminability across contraction intensities, we further present an sEMG fatigue dataset collected from 30 participants at four levels of maximum voluntary contraction (MVC) and benchmark representative machine learning and deep learning methods. We evaluate performance offline under leave-one-subject-out (LOSO) and within-subject protocols and use the SWIFTIES dataset for external validation. Results support FatigueFormer's effectiveness across the examined contraction intensities and both datasets, while indicating that fatigue-related cues vary across MVC levels. An exploratory adaptation analysis shows higher FatigueFormer scores with more labeled target-user data on complete target recordings, which include calibration samples.
\end{abstract}

\section{Introduction}
Surface electromyography (sEMG) supports non-invasive neuromotor interfaces for myoelectric prosthetic control and human--computer interaction, including gesture recognition and text entry~\cite{Kaifosh2025GenericNeuromotor,Salter2024Emg2pose,Sivakumar2024Emg2qwerty}. Recorded at the skin surface, sEMG signals reflect superimposed motor unit action potentials (MUAPs) and carry information about motor commands from the central nervous system~\cite{Kaifosh2025GenericNeuromotor}. However, sustained activity can induce muscle fatigue through neural and muscular processes that affect force-generating capacity~\cite{Musclefatigue} and alter sEMG amplitude, spectrum and complexity~\cite{Electromyography,PHINYOMARK20127420}. To reduce movement-related confounding, we assess fatigue during sustained isometric elbow flexion using sEMG recorded from the biceps and triceps.

Traditional sEMG analyses have shown that some handcrafted descriptors increase or decrease as fatigue develops~\cite{Ming2014FatigueTrends,KARTHICK201642}. Joint analysis of the EMG spectrum and amplitude (JASA) combines the directions of amplitude and spectral changes to distinguish fatigue-related changes from changes in force production~\cite{Luttmann2000JASA}. Learning-based methods use convolutional and temporal architectures to model raw sEMG or sequences of handcrafted descriptors~\cite{Yin2026WristFatigue,Zhang2021MFFNet,Yu2024}. The representative architectures cited above can learn temporal changes in sEMG, but do not explicitly use the increasing and decreasing descriptor trends reported in physiological studies to guide signal encoding. On the other hand, sEMG descriptors vary with contraction intensity~\cite{Arjunan2014FatigueFeatures} and across individuals. Information that distinguishes fatigue states within one participant may not remain equally informative for another. Some fatigue-recognition methods have been evaluated at only one or a few contraction intensities~\cite{Zhang2024}, leaving their performance at other loads unclear. Evaluation across a broader range of contraction intensities, under both within-subject and cross-subject protocols, is therefore needed to examine recognition performance and the contributions of individual features and model components.

To address these limitations, we propose FatigueFormer to recognize three fatigue states defined by ratings of perceived exertion (RPE). Based on how handcrafted descriptors change across fatigue states, we assign those with significantly positive or negative slopes ($p<0.05$) to increasing-trend (INC) or decreasing-trend (DEC) streams, respectively. We build temporal and static branches that share this INC/DEC grouping. The temporal branch models inter-window context, while the static branch captures descriptor interactions within the target window. Within each branch, self-attention models relationships within each trend stream, and cross-attention exchanges information between the INC and DEC streams. The resulting temporal and static representations are fused for fatigue-state prediction.

We evaluate FatigueFormer on our four-intensity sEMG dataset and the independent SWIFTIES dataset under leave-one-subject-out (LOSO) and within-subject protocols. FatigueFormer outperforms most evaluated baselines under LOSO and remains competitive within subject. We further examine target-participant adaptation using labeled target data.

Our main contributions are as follows:
\begin{itemize}
\item We propose FatigueFormer, which incorporates fatigue-related trend grouping into a dual-branch architecture to jointly model inter-window context and within-window descriptor interactions.
\item We introduce a 30-participant sEMG fatigue dataset covering four contraction intensities (20--80\% of maximum voluntary contraction, MVC), and establish benchmarks under LOSO and within-subject protocols.
\item We identify a stable descriptor core shared across cohorts, characterize intensity-dependent feature selection, and assess component contributions across contraction intensities and evaluation protocols.
\end{itemize}

\section{Related Work}
\paragraph{sEMG decoding and fatigue assessment}
EV-Action combines EMG with visual and skeletal modalities for action recognition~\cite{Wang_2020}, while Eddy et al. study cross-user sEMG gesture recognition~\cite{Eddy2024}. Kaifosh et al. improve wrist-sEMG decoding for unseen users by training on more participants and demonstrate benefits from target-user fine-tuning for handwriting~\cite{Kaifosh2025GenericNeuromotor}. For fatigue assessment, Cerqueira et al. pair upper-limb sEMG with subjective ratings during dynamic movements~\cite{Cerqueira2024}. We study RPE-derived fatigue during sustained isometric upper-limb exertion, examining both cross-subject recognition and target-user personalization.

\paragraph{Fatigue representations and feature interactions}
Amplitude, spectral and complexity measures describe complementary fatigue-related changes in sEMG~\cite{DeLuca1997,PHINYOMARK20127420}. During fatiguing isometric contractions, mean and median frequencies decrease while the root mean square (RMS) of low-frequency wavelet coefficients increases~\cite{Ming2014FatigueTrends}. Time--frequency trends also occur during dynamic biceps contractions~\cite{KARTHICK201642}. JASA combines amplitude and spectral trends through quadrant rules to distinguish fatigue-related changes from changes in force production~\cite{Luttmann2000JASA}.

Handcrafted features support conventional classifiers~\cite{article7} and temporal models, including wavelet-packet LSTM for cycling~\cite{Wang2021LSTMFatigue} and CNN--LSTM operating on multi-domain feature sequences~\cite{Yin2026WristFatigue}. Other fatigue models use attention-based LSTM~\cite{DANG2023112} or CNN--LSTM--Transformer architectures~\cite{Yu2024}. MFFNet combines convolutional and bidirectional recurrent representations through attention~\cite{Zhang2021MFFNet}.

Beyond temporal modeling, general-purpose feature-interaction methods offer tools for learning relationships among sEMG descriptors. AutoInt uses self-attention to model feature interactions~\cite{Song_2019}, while feature tokenization embeds individual measurements for Transformer-based modeling~\cite{gorishniy2023revisitingdeeplearningmodels,vaswani2023attentionneed}. In gesture recognition, TraHGR combines complementary signal-patch pathways, with FNet retaining both spatial and temporal information~\cite{10285513}. FatigueFormer builds on temporal and feature-interaction modeling by organizing descriptors into physiologically motivated INC/DEC groups within each branch.

\paragraph{Contraction intensity and evaluation}
Contraction intensity affects fatigue-related sEMG patterns~\cite{Arjunan2014FatigueFeatures,article3}. MACNet evaluates within-subject and cross-subject recognition and uses GradCAM to examine channel and temporal contributions at 50\% MVC~\cite{Zhang2024}. This single-intensity setting leaves variation in feature and component contributions across loads unresolved. We examine these contributions across four MVC levels under LOSO and within-subject protocols. Independent evaluation on SWIFTIES further tests whether the observed patterns recur at low-to-moderate intensities~\cite{OSullivan2024SWIFTIES}.

\section{FatigueFormer}
\label{sec:method}
\begin{figure*}[t]
\centering
\includegraphics[width=\textwidth]{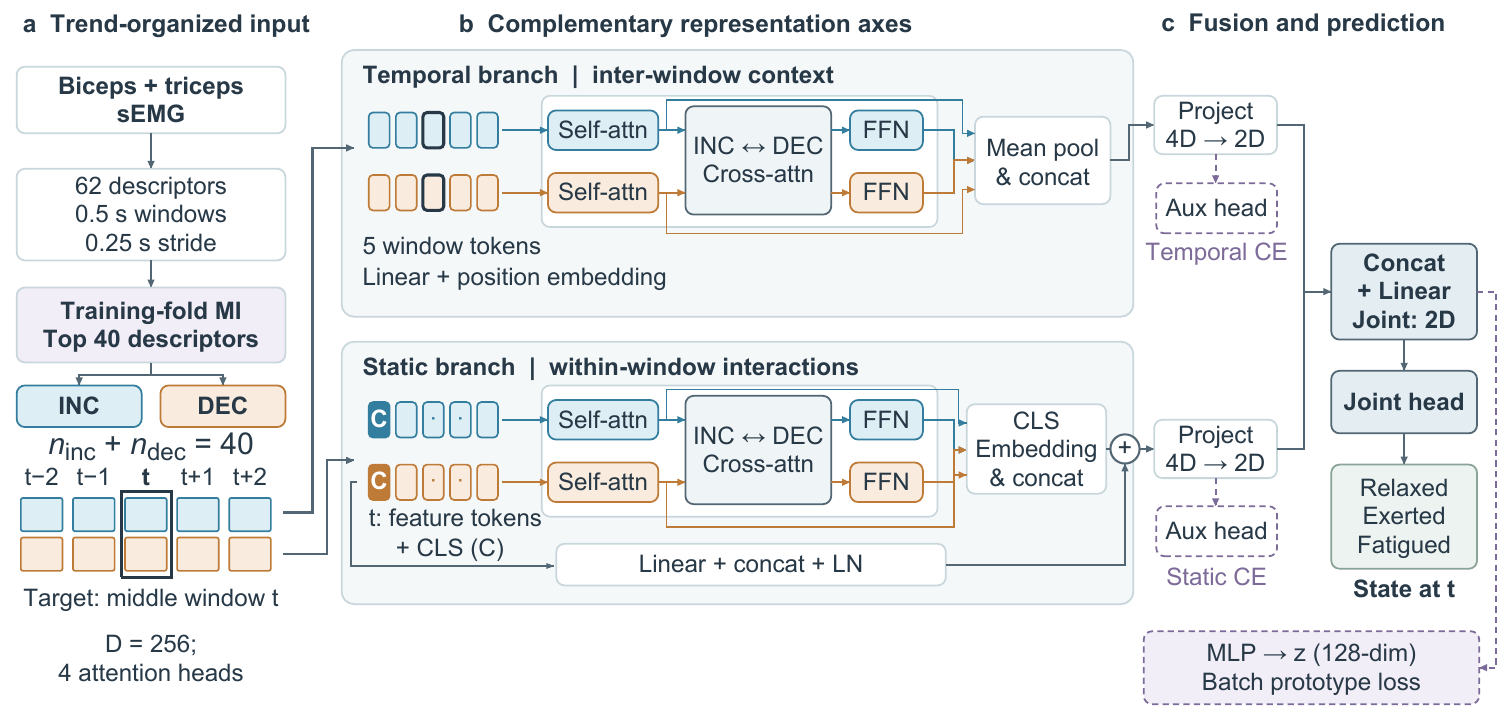}
\caption{FatigueFormer overview. (a) Training-fold MI selects 40 descriptors across the increasing (INC) and decreasing (DEC) streams; $t$ marks the target window. (b) The temporal branch treats five windows as tokens, whereas the static branch tokenizes features at $t$ and adds a linear path. Each branch combines self-attended and cross-updated representations from both streams. (c) Projected branch representations are concatenated and mapped to a joint representation for three-state prediction. Dashed routes are used only during training. Purple-shaded modules (MI selection and prototype regularization) are included only in the full configuration; the base configuration uses all 62 descriptors. Some residual and normalization operations are omitted for clarity. LN denotes layer normalization.}
\label{fig:framework}
\end{figure*}

FatigueFormer combines inter-window context with within-window feature interactions to recognize fatigue states during sustained upper-limb exertion (Fig.~\ref{fig:framework}). The temporal branch embeds each window's descriptor vector as a window token. The static branch uses feature tokenization to map individual descriptors within the target window to feature tokens. Both branches organize descriptors into increasing (INC) and decreasing (DEC) trend streams and learn interactions within and between them. Their representations are fused for three-state prediction. The optional full configuration adds mutual information (MI) feature selection and a prototype loss.

\subsection{Input representation and trend organization}
\label{sec:input}
We extract 31 descriptors per muscle from biceps and triceps sEMG using 0.5\,s windows with a 0.25\,s stride. At window $t$, the handcrafted descriptors from both muscles form a feature vector $x_t\in\mathbb{R}^{62}$. Definitions and trend assignments are given in Supplementary Section~S2, Table~II.

The ordinal label $y_t\in\{0,1,2\}$ denotes \emph{relaxed}, \emph{exerted} and \emph{fatigued} states, respectively, corresponding to Borg RPE ranges 6--10, 11--15 and 16--20~\cite{Borg1998}. These labels describe perceived exertion during the prescribed contraction.

Guided by physiological evidence of fatigue-related descriptor trends~\cite{Arjunan2014FatigueFeatures,Ming2014FatigueTrends}, we evaluate candidate handcrafted descriptors using labeled observations pooled across participants and MVC levels. We regress each candidate descriptor against the ordinal fatigue labels and retain 31 descriptors with significant slopes ($p<0.05$), assigning the 15 with positive slopes to INC and the 16 with negative slopes to DEC. Applying this grouping to both muscle channels yields 30 INC and 32 DEC descriptors. The descriptor set and trend assignments remain fixed across experiments.

In the full configuration, we further select the $K=40$ descriptors with the highest MI with the target state within each training fold:
\begin{equation}
\mathcal{S}=\operatorname{TopK}_{j}\,I(X_j;Y),\qquad K=40.
\end{equation}
Here, $X_j$ and $Y$ denote the random variables corresponding to descriptor $j$ and the state label, and $\mathcal{S}$ contains the selected descriptor indices. MI is estimated using only the target-window features and labels from the training set. The selected indices are applied unchanged to the validation and test sets, preserving the fixed INC/DEC assignments. Thus, $n_{\mathrm{inc}}+n_{\mathrm{dec}}=40$ in the full configuration, where $n_{\mathrm{inc}}$ and $n_{\mathrm{dec}}$ denote the numbers of selected descriptors in the INC and DEC streams, respectively. The base configuration skips MI selection and retains all 62 descriptors. The temporal branch uses five consecutive windows centered on $t$; the static branch uses only window $t$.

\subsection{Inter-window temporal modeling}
The temporal branch represents the local evolution of muscular activity by treating each window as one token. For stream $s\in\{\mathrm{inc},\mathrm{dec}\}$, its five-window input is $X_s\in\mathbb{R}^{5\times n_s}$. Stream-specific linear projections embed the descriptors, and a shared learned positional encoding distinguishes their temporal order:
\begin{equation}
H_s^{(0)}=X_s W_s+P.
\end{equation}
Here, $W_s\in\mathbb{R}^{n_s\times D}$ is a stream-specific learnable projection matrix, and $P\in\mathbb{R}^{5\times D}$ is a learnable positional embedding shared by both streams. $H_s^{(0)}\in\mathbb{R}^{5\times D}$ denotes the initial window-token sequence with embedding dimension $D$.

The interaction blocks progressively combine information within and between the trend streams. In each block, self-attention first models inter-window relationships within INC and DEC separately. Bidirectional cross-attention then uses each stream as queries and the opposite self-attended stream as keys and values, allowing the two temporal patterns to inform each other. A feed-forward network completes the update; the blocks use layer normalization and residual connections.

The temporal representation retains both within-stream and cross-stream information. Let $\bar h^s$ denote the mean-pooled self-attention output of the final block, and $\bar m^s$ its mean-pooled output after cross-attention and the feed-forward update. Their concatenation gives
\begin{equation}
f_T=[\bar h^{\mathrm{inc}};\bar m^{\mathrm{inc}};
       \bar h^{\mathrm{dec}};\bar m^{\mathrm{dec}}]\in\mathbb{R}^{4D}.
\end{equation}
Thus, $f_T$ summarizes the surrounding sequence while preserving separate summaries of each trend before and after cross-stream interaction.

\subsection{Within-window feature interactions}
Temporal pooling summarizes the surrounding context but can attenuate the configuration of the target window itself. The static branch therefore models relationships among descriptors at $t$, using features rather than windows as tokens. Each selected scalar is mapped to a feature-specific embedding:
\begin{equation}
t_j^s=x_j^s a_j^s+b_j^s,\qquad a_j^s,b_j^s\in\mathbb{R}^{D},
\end{equation}
where $x_j^s$ denotes scalar descriptor $j\in\{0,\ldots,n_s-1\}$ in stream $s$ at the target window, $a_j^s$ and $b_j^s$ are feature-specific learnable weight and bias vectors, and $t_j^s\in\mathbb{R}^{D}$ is the resulting feature token.

The tokenizer parameters encode feature identity, so no positional encoding is added. We prepend a separate learnable classification (CLS) token $c_s\in\mathbb{R}^{D}$ to each stream (marked C in Fig.~\ref{fig:framework}):
\begin{equation}
T_s^{(0)}=[c_s,t_0^s,\ldots,t_{n_s-1}^s]\in\mathbb{R}^{(n_s+1)\times D}.
\label{eq:cls-input}
\end{equation}
Here, tokens are stacked along the sequence dimension. Following FT-Transformer~\cite{gorishniy2023revisitingdeeplearningmodels}, a learnable CLS token aggregates feature information into a fixed-dimensional representation for classification.

We feed the CLS-augmented INC and DEC sequences into the interaction block, which applies self-attention within each trend, followed by bidirectional cross-attention between trends and a feed-forward update. Each CLS embedding thus summarizes both within-trend and cross-trend descriptor information.

For each stream, we retain the $D$-dimensional hidden state of its CLS token at two stages: $h_{\mathrm{CLS}}^s$ after the self-attention update, and $m_{\mathrm{CLS}}^s$ after cross-attention and the feed-forward update. As in the temporal branch, their concatenation gives
\begin{equation}
f_{\mathrm{seq}}=[h_{\mathrm{CLS}}^{\mathrm{inc}};m_{\mathrm{CLS}}^{\mathrm{inc}};
                  h_{\mathrm{CLS}}^{\mathrm{dec}};m_{\mathrm{CLS}}^{\mathrm{dec}}]\in\mathbb{R}^{4D}.
\label{eq:static-readout}
\end{equation}

A parallel residual path preserves the selected target-window descriptors $x_t^s\in\mathbb{R}^{n_s}$. For each stream $s\in\{\mathrm{inc},\mathrm{dec}\}$, a learnable $\operatorname{Linear}_s$ projection maps them to $2D$ dimensions. The residual and static representations are
\begin{equation}
\begin{aligned}
r_s&=\operatorname{LN}\!\left(\operatorname{Linear}_s(x_t^s)\right),\\
f_S&=\operatorname{LN}(f_{\mathrm{seq}})
     +\operatorname{LN}([r_{\mathrm{inc}};r_{\mathrm{dec}}]).
\end{aligned}
\end{equation}
Here, $r_s\in\mathbb{R}^{2D}$, and each normalization layer has independent parameters.

The temporal and static branches have independent parameters. Cross-attention operates between INC and DEC within each branch; fusion subsequently combines the two branches. This separation allows component analysis to distinguish the contributions of inter-window context, within-window interactions and cross-trend exchange while retaining the same input descriptor set.

\subsection{Representation fusion and training}
Linear projections map $f_T$ and $f_S$ to $2D$ dimensions; their concatenation is mapped to a joint $2D$ representation for fatigue-state prediction. The joint head and two auxiliary heads predict probabilities over the three fatigue states from the fused, temporal and static representations, respectively. We denote their class-weighted cross-entropy losses with label smoothing by $\mathcal{L}_{\mathrm{joint}}$, $\mathcal{L}_{T}$ and $\mathcal{L}_{S}$. Class weights are inversely proportional to class frequencies in the training set, and the label-smoothing factor is 0.1.

In the full configuration, we additionally regularize the fused representation with class prototypes. A two-layer projection head maps the fused representation to a normalized 128-dimensional embedding $z_i$. Each prototype $p_c$ is the normalized mean of embeddings with label $c$ in the current mini-batch. The prototype objective encourages each embedding to align with its class anchor relative to the other represented classes:
\begin{equation}
\mathcal{L}_{\mathrm{proto}}=-\frac{1}{N}\sum_i
\log\frac{\exp(z_i^\top p_{y_i}/\tau)}
{\sum_c\exp(z_i^\top p_c/\tau)}.
\end{equation}
Here, $N$ is the mini-batch size, $y_i$ is the target-window label, and $\tau=0.1$ is the temperature. The denominator sums over classes present in the batch. The total loss is
\begin{equation}
\mathcal{L}=\mathcal{L}_{\mathrm{joint}}+
\mathcal{L}_{T}+\mathcal{L}_{S}+0.5\mathcal{L}_{\mathrm{proto}}.
\end{equation}
Auxiliary heads and prototype regularization are used only during training; inference uses the joint head.

We use embedding width $D=256$, four attention heads, feed-forward width $4D$, and dropout 0.5 in the classification heads. Architectural settings remain fixed across protocols, contraction intensities and cohorts. The base variant retains the same backbone and three-head classification supervision, uses all 62 descriptors, and omits MI selection and the prototype objective. Further implementation details are provided in Supplementary Section~S3.

\section{Datasets}
\label{sec:datasets}
\begin{table}[t]
\centering\small
\caption{Cohorts and pooled window counts. Primary-cohort counts precede the exclusions needed for valid within-subject folds.}
\label{tab:data}
\begin{tabular}{lrrr}
\toprule
Cohort & Participants & MVC (\%) & Windows\\
\midrule
Primary & 30 (15 M/15 F) & 20 & 32,597\\
 & & 40 & 9,124\\
 & & 60 & 4,138\\
 & & 80 & 2,818\\
\midrule
SWIFTIES & 32 (17 M/15 F) & 25 & 22,795\\
 & & 45 & 12,082\\
\bottomrule
\end{tabular}
\end{table}

\subsection{Primary dataset}
\begin{figure}[t]
\centering
\includegraphics[width=0.80\columnwidth]{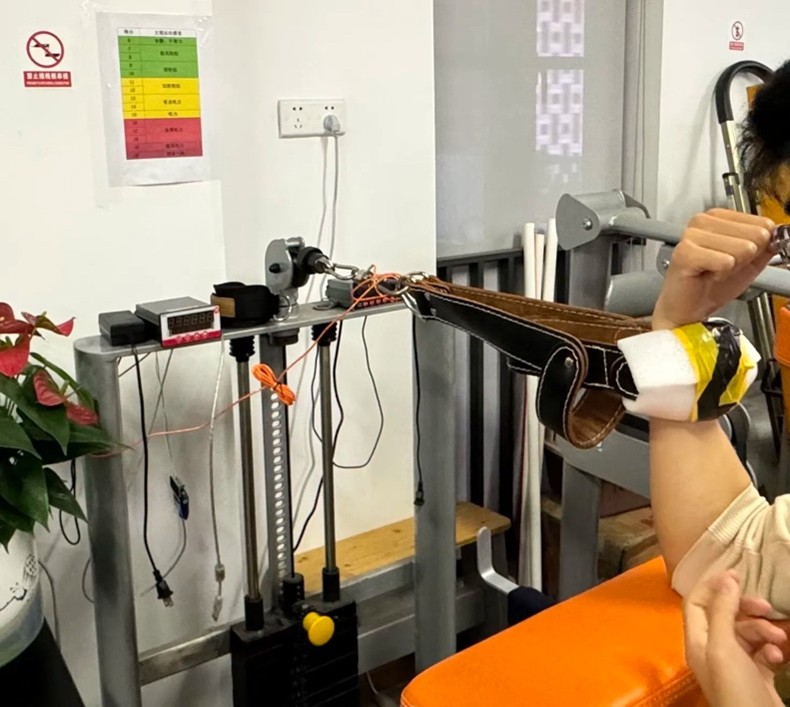}
\caption{Force-feedback apparatus and arm positioning during sustained isometric elbow flexion. The biceps and triceps electrodes are outside the illustrated view.}
\label{fig:acquisition}
\end{figure}

We collected an sEMG dataset during sustained isometric elbow-flexion contractions to study fatigue-state recognition across contraction intensities. Thirty healthy adults, 15 male and 15 female, performed the task at 20\%, 40\%, 60\% and 80\% of their maximum voluntary contraction (MVC). Target forces were set using each participant's MVC measured before the fatigue trials. Participants were seated with the upper arm stabilized and used real-time visual force feedback to maintain the prescribed contraction (Fig.~\ref{fig:acquisition}). Each trial ended when the participant could no longer maintain at least 90\% of the target force. All participants provided written informed consent before acquisition.

After cleaning the skin with alcohol, we placed bipolar electrodes over the biceps brachii and triceps brachii muscle bellies and secured them with gauze. Both channels were recorded at 2,000\,Hz from contraction onset to task failure. Signals were band-pass filtered at 20--450\,Hz and notch filtered at 50\,Hz.

During each contraction, participants reported perceived exertion in real time using Borg's RPE scale (6--20). The acquisition device timestamped each report for alignment with the sEMG recordings. These reports provided the three fatigue-state labels defined in Section~\ref{sec:method}.

Using the windowing procedure described in Section~\ref{sec:input}, we obtained 48,677 windows across the four contraction intensities (Table~\ref{tab:data}). Pooled counts decrease from 32,597 at 20\% MVC to 2,818 at 80\% MVC, reflecting the shorter sustainable recording durations at higher intensities. Detailed acquisition procedures and class-specific window counts are provided in Supplementary Section~S1.

\subsection{SWIFTIES dataset}
We use the SWIFTIES dataset~\cite{OSullivan2024SWIFTIES} for external validation. It contains recordings from 32 healthy adults, 17 male and 15 female, across shoulder and elbow tasks. We select its isometric elbow-flexion conditions at 25\% and 45\% MVC, covering a low-to-moderate contraction-intensity range. The target load is approximated using the closest available dumbbell mass, whereas the primary dataset uses real-time force feedback. For analysis, we use the biceps and triceps channels and the same 62-descriptor input format.

Borg RPE scores were recorded every 15\,s at 25\% MVC and every 10\,s at 45\% MVC. We interpolate the discrete reports over time to estimate an RPE trajectory aligned with the sEMG windows. The resulting ratings are then mapped to the same three fatigue states defined in Section~\ref{sec:method}.

\section{Experiments}
\label{sec:evaluation}
\label{sec:results}
\subsection{Experimental setup}
\paragraph{Baselines and implementation}
We compare against AutoInt~\cite{Song_2019}, RBF-SVM~\cite{article8}, CNN--SVM~\cite{10211798}, a CNN--LSTM--Transformer (CLT)~\cite{Yu2024}, and MACNet~\cite{Zhang2024}. AutoInt and RBF-SVM use the same 62-descriptor pool supplied to FatigueFormer before MI selection; CNN--SVM, CLT and MACNet use two-channel raw sEMG. We evaluate both base and full FatigueFormer. Two additional controls are trained from scratch and evaluated under primary-cohort LOSO. Feature-LT62 feeds all 62 handcrafted descriptors to a bidirectional LSTM--Transformer, without a CNN front end or MI selection. Full--Swap50 partially exchanges INC/DEC assignments after MI selection, preserving the selected descriptors and stream sizes. For comparison, we adapt each model to the dataset-specific inputs and train it from scratch. For the primary cohort, neural models use Adam~\cite{kingma2017adammethodstochasticoptimization} and validation-accuracy early stopping. Input construction, protocol-specific scaling and hyperparameters are detailed in Supplementary Section~S3.

\paragraph{Evaluation protocols}
Models are trained and evaluated separately within each dataset and MVC level. LOSO holds out one participant for testing, with that participant's samples excluded from model fitting. Deep learning models reserve one additional participant for validation. This assesses recognition for a participant absent from the training set. Within-subject evaluation divides each participant's samples into subsets called folds, using each fold in turn for testing and the remaining folds for training and validation. This assesses fatigue recognition for a participant represented in the training data. For primary-cohort FatigueFormer, we first construct target samples with complete five-window context. We then form up to five stratified folds according to class support and omit splits whose training sets lack any fatigue class. Samples are constructed before fold assignment, so folds may share signal samples or adjacent context. Split details are in Supplementary Section~S3; participant counts are in Sections~S5 and S8.

\begin{table*}[t]
\centering\small
\setlength{\tabcolsep}{3pt}
\caption{Macro-F1 (\%) by dataset, protocol and MVC. WS: within subject. Bold and underlined values indicate the highest and second-highest reported means per column, respectively. Primary-cohort WS RBF-SVM uses $n=30/29/30/26$ across MVC levels. SWIFTIES uses $n=32$, seed 0; external-model comparisons are descriptive, without paired tests. Primary-cohort statistical results appear in Supplementary Tables~VIII--IX; evaluation and implementation details are in Section~\ref{sec:evaluation} and Supplementary Section~S3.}
\label{tab:performance}
\begin{tabular}{@{}l*{8}{r}@{\hspace{9pt}}*{4}{r}@{}}
\toprule
&\multicolumn{8}{c}{Primary cohort}&\multicolumn{4}{c}{SWIFTIES}\\
\cmidrule(lr){2-9}\cmidrule(lr){10-13}
&\multicolumn{4}{c}{WS}&\multicolumn{4}{c}{LOSO}&\multicolumn{2}{c}{WS}&\multicolumn{2}{c}{LOSO}\\
\cmidrule(lr){2-5}\cmidrule(lr){6-9}\cmidrule(lr){10-11}\cmidrule(lr){12-13}
Method &20\%&40\%&60\%&80\%&20\%&40\%&60\%&80\%&25\%&45\%&25\%&45\%\\
\midrule
AutoInt &86.31&85.56&78.62&71.49&41.21&51.63&43.79&36.49&72.71&69.90&54.37&60.66\\
RBF-SVM &89.61&\textbf{94.62}&\textbf{92.59}&\textbf{92.75}&42.01&51.47&46.13&40.52&77.23&78.02&51.39&57.25\\
CNN--SVM &54.89&53.50&43.47&31.04&34.60&37.57&32.92&31.04&50.44&43.06&38.59&39.70\\
CLT &33.24&24.23&18.61&17.13&24.15&26.52&26.81&23.80&\textbf{94.48}&\textbf{90.18}&42.02&39.90\\
MACNet &24.62&22.13&20.03&19.28&17.81&10.88&21.82&25.68&67.65&67.82&44.04&42.26\\
\midrule
FatigueFormer base &\textbf{93.52}&\underline{92.10}&81.79&87.16&\underline{53.59}&\textbf{72.52}&\textbf{59.55}&\underline{55.24}&\underline{89.00}&\underline{85.09}&\textbf{60.86}&\textbf{67.90}\\
FatigueFormer full &\underline{92.97}&91.95&\underline{82.05}&\underline{88.81}&\textbf{55.41}&\underline{72.34}&\underline{57.56}&\textbf{60.15}&87.57&83.68&\underline{58.02}&\underline{66.62}\\
\bottomrule
\end{tabular}
\end{table*}

\paragraph{Evaluation metrics}
Macro-F1 is computed for each test split as
\begin{equation}
\mathrm{Macro\text{-}F1}
=\frac{1}{|\mathcal{C}|}\sum_{c\in\mathcal{C}}
\frac{2\mathrm{TP}_c}{2\mathrm{TP}_c+\mathrm{FP}_c+\mathrm{FN}_c},
\label{eq:macro-f1}
\end{equation}
where $\mathcal{C}$ contains the classes present in the test labels or predictions, and $\mathrm{TP}_c$, $\mathrm{FP}_c$ and $\mathrm{FN}_c$ denote class-specific true-positive, false-positive and false-negative counts. Scores are averaged over available fold--seed combinations within each participant and then equally across participants.

We estimate 95\% confidence intervals using participant-level bias-corrected and accelerated (BCa) bootstrap resampling. For inferential comparisons, we use paired one-sided Wilcoxon tests with Benjamini--Hochberg false discovery rate (FDR) correction within each MVC (Supplementary Section~S3). Corrected $q<0.05$ indicates statistical significance. Matched-pairs rank-biserial $r\in[-1,1]$ describes effect size: positive values favor full FatigueFormer, negative values favor the comparator, and larger $|r|$ indicates a stronger rank-based effect. We interpret $r$ alongside the corresponding macro-F1 difference. Table~\ref{tab:performance} summarizes mean macro-F1, and Fig.~\ref{fig:ablation} shows changes relative to the full model. Detailed statistics are reported in Supplementary Tables~VIII--XI for the primary cohort and Table~VI for the SWIFTIES ablations.

\subsection{Model comparison}
FatigueFormer achieves higher macro-F1 than most evaluated external models across contraction intensities under both LOSO and within-subject protocols (Table~\ref{tab:performance}).

\paragraph{LOSO evaluation}
In the primary dataset, full FatigueFormer achieves 55.41--72.34\% macro-F1 and exceeds the five external baselines in Table~\ref{tab:performance} at each intensity after FDR correction. Its margins over the strongest of these baselines are 13.40, 20.71, 11.43 and 19.63 percentage points at 20\%, 40\%, 60\% and 80\% MVC, respectively. In SWIFTIES, full-model macro-F1 increases from 58.02\% at 25\% MVC to 66.62\% at 45\%, exceeding the strongest external comparator, AutoInt, by 3.65 and 5.96 points. Among these primary-cohort comparisons, the narrowest margin, against RBF-SVM at 60\% MVC, remains significant ($q=0.014$). Comparisons with CNN--SVM, CLT and MACNet yield $r\geq0.85$ at all four primary-cohort intensities.

Among the evaluated intensities, full FatigueFormer achieves its highest LOSO macro-F1 at 40\% MVC in the primary dataset and 45\% MVC in SWIFTIES. Most external baselines also attain their best LOSO scores at these respective intensities. Both conditions contain fewer windows than their corresponding lower-intensity conditions (Table~\ref{tab:data}), indicating that recognition performance does not simply improve with sample count and that recognition difficulty varies across MVC levels. This recurring pattern across cohorts supports multi-MVC data collection and separate evaluation at each contraction intensity.

\begin{figure}[t]
\centering
\includegraphics[width=\columnwidth]{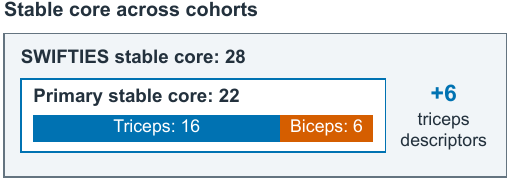}
\caption{Stable descriptor cores in the primary and SWIFTIES datasets. The primary core (22 descriptors: 16 triceps and six biceps) is a strict subset of the SWIFTIES core (28), which contains six additional triceps descriptors. A core descriptor is selected in every LOSO training fold at every evaluated MVC level within its cohort.}
\label{fig:features}
\end{figure}

\begin{figure*}[t]
\centering
\includegraphics[width=\textwidth]{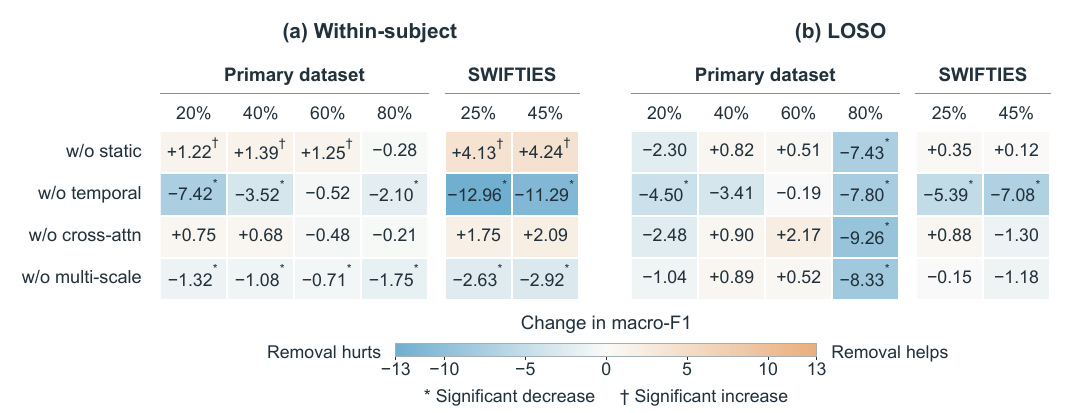}
\caption{Component ablations in the primary dataset (20--80\% MVC) and SWIFTIES (25\% and 45\% MVC), under (a) within-subject and (b) LOSO evaluation. Cells show changes in mean macro-F1 relative to full FatigueFormer, in percentage points (variant minus full), using a shared color scale. Negative values favor retaining the component. $*$ marks an FDR-significant decrease ($q<0.05$); $\dagger$ marks a significant increase under the reverse test. Unmarked cells show no reported significant effect. Full-model scores are in Table~\ref{tab:performance}; detailed ablation means and paired tests are in Supplementary Table~VI (SWIFTIES) and Tables~X--XI (primary dataset).}
\label{fig:ablation}
\end{figure*}

\paragraph{Within-subject evaluation}
In the primary dataset, RBF-SVM achieves the highest within-subject macro-F1 at 40--80\% MVC, scoring 92.59--94.62\%. FatigueFormer ranks second in either its full or base configuration at these intensities, while its two configurations rank first and second at 20\% MVC. In SWIFTIES, CLT leads at 94.48\% and 90.18\% for 25\% and 45\% MVC, respectively, with base FatigueFormer ranking second at both intensities. FatigueFormer therefore remains competitive within subject across both datasets.

High within-subject performance does not necessarily persist under LOSO. In the primary dataset, RBF-SVM's reported macro-F1 falls from 89.61--94.62\% within subject to 40.52--51.47\% under LOSO, corresponding to reductions of 43.15--52.23 percentage points across intensities. Similarly, CLT in SWIFTIES falls from 94.48\% and 90.18\% within subject to 42.02\% and 39.90\% under LOSO at 25\% and 45\% MVC, respectively, corresponding to drops of 52.46 and 50.28 percentage points. These differences in model performance suggest that patterns distinguishing fatigue states within one participant may not remain equally discriminative across participants. LOSO therefore places greater demands on representations that remain informative \mbox{across participants}.

\paragraph{Full and base configurations}
Full and base FatigueFormer have different strengths across intensities and evaluation protocols, while both remain competitive. Without MI selection or prototype regularization, base still outperforms the five external baselines in Table~\ref{tab:performance} under LOSO in both datasets, showing that the shared trend-organized, dual-branch architecture already supports strong recognition. Full achieves higher scores under both protocols at 80\% MVC in the primary dataset, where fewer windows are available. One possible explanation is that selecting informative descriptors and encouraging compact class representations help when training data are limited. This comparison concerns the complete configurations, so it does not isolate the effect of either addition.

\paragraph{Feature representations and trend grouping}
Under primary-cohort LOSO, Feature-LT62 outperforms CLT at all four contraction intensities, and FatigueFormer base further outperforms Feature-LT62 (Table~\ref{tab:feature-lt}). The improvement with handcrafted descriptor sequences supports feature-based temporal modeling for this task. The additional gains of FatigueFormer base, which also retains all 62 descriptors, suggest that how these descriptors are organized and modeled matters. Its temporal and static branches may capture complementary fatigue information through inter-window context and within-window descriptor interactions, with INC/DEC grouping guiding the modeling of both.

Full FatigueFormer outperforms Full--Swap50 at three of the four intensities, with the largest advantage at 80\% MVC; the partial-exchange control performs slightly better at 40\% MVC. This control retains the full architecture, MI-selected descriptors and stream sizes while partially disrupting trend assignments. One possible explanation is that grouping descriptors with similar fatigue-related trends helps model their relationships within each stream, while cross-attention combines complementary information across trends. This single fixed-exchange control suggests that the contribution of trend organization varies with contraction intensity.

\begin{table}[t]
\centering\small
\setlength{\tabcolsep}{4pt}
\caption{Primary-cohort LOSO macro-F1 (\%). Feature-LT62 uses all 62 descriptors. Full--Swap50 partially exchanges INC/DEC assignments after MI selection, preserving stream sizes. Bold/underlined: highest/second-highest per column. The two control comparisons are descriptive.}
\label{tab:feature-lt}
\begin{tabular}{@{}lrrrr@{}}
\toprule
Model &20\%&40\%&60\%&80\%\\
\midrule
CLT &24.15&26.52&26.81&23.80\\
Feature-LT62 &53.42&68.41&\underline{58.83}&51.85\\
\midrule
Full--Swap50 &\underline{54.01}&\textbf{72.77}&56.97&52.43\\
FatigueFormer base &53.59&\underline{72.52}&\textbf{59.55}&\underline{55.24}\\
FatigueFormer full &\textbf{55.41}&72.34&57.56&\textbf{60.15}\\
\bottomrule
\end{tabular}
\end{table}

\subsection{Personalization to new participants}
\label{sec:personalization}
Prior work on generic sEMG interfaces reported power-law reductions in decoding error on unseen participants as training cohorts and model capacity increased~\cite{Kaifosh2025GenericNeuromotor}. In handwriting decoding, fine-tuning with target-user data provided an effective alternative to further expanding the training cohort. Motivated by these findings, we examine personalization of full FatigueFormer in the primary dataset. Starting from the corresponding LOSO checkpoint trained on other participants, we fine-tune the model for ten epochs using 20\% or 40\% of the target participant's fatigue-labeled data, with all three fatigue states represented in each adaptation set. Both settings are evaluated on each model's complete original target test set.

Table~\ref{tab:personalization} provides a descriptive comparison of the three models under both adaptation budgets. At 20--60\% MVC, full FatigueFormer ranks first, followed by descriptor-based RBF-SVM and raw-signal CNN--SVM; CNN--SVM scores slightly higher than full FatigueFormer at 80\% MVC. This pattern suggests that handcrafted descriptors remain useful for personalized fatigue recognition. Full FatigueFormer also outperforms RBF-SVM at every intensity under both budgets, showing that its advantage persists when target-user labels are available. Its trend-organized representations learned from other participants may provide a useful starting point for adapting to individual fatigue patterns.

\begin{table}[t]
\centering\small
\setlength{\tabcolsep}{2.5pt}
\caption{Target-participant adaptation: macro-F1 (\%) on each model's complete original TEST, including calibration samples. Budgets are nominal and sample supports differ across models; comparisons are descriptive. Bold marks the largest reported value per budget and MVC.}
\label{tab:personalization}
\begin{tabular}{@{}llrrrr@{}}
\toprule
\multirow{2}{*}{Budget} &\multirow{2}{*}{Model} &\multicolumn{4}{c}{MVC (\%)}\\
\cmidrule(l){3-6}
&&20&40&60&80\\
\midrule
\multirow{3}{*}{20\%} &RBF-SVM &57.32&65.89&62.62&55.10\\
&CNN--SVM &54.62&54.59&51.52&\textbf{63.92}\\
&FatigueFormer full &\textbf{68.72}&\textbf{80.41}&\textbf{69.82}&62.90\\
\midrule
\multirow{3}{*}{40\%} &RBF-SVM &65.39&73.27&69.26&63.55\\
&CNN--SVM &65.10&62.50&53.99&\textbf{68.62}\\
&FatigueFormer full &\textbf{77.69}&\textbf{83.70}&\textbf{73.84}&68.43\\
\bottomrule
\end{tabular}
\end{table}

\subsection{Feature analysis across contraction intensities}

Under primary-cohort LOSO, 22 of the 62 descriptors are selected in all 30 training folds at each of the four contraction intensities. We refer to this set as the \emph{stable core} (Fig.~\ref{fig:features}). It contains 16 triceps and six biceps descriptors, spanning frequency, waveform structure, complexity, recurrence, autocorrelation and wavelet ratios. The repeated selection of triceps descriptors may reflect fatigue-related changes in antagonist muscle activity during sustained elbow flexion, consistent with reports of antagonist coactivation~\cite{Levenez2008Coactivation}.

In SWIFTIES, 28 descriptors are selected in all 32 LOSO training folds at both 25\% and 45\% MVC. This core contains all 22 primary-core descriptors and six additional triceps descriptors (Fig.~\ref{fig:features}). The narrower MVC range may partly explain why more descriptors meet the across-intensity consistency requirement. The retention of all 22 descriptors in the primary dataset's stable core supports reproducibility across independent cohorts and acquisition conditions within the evaluated low-to-moderate intensity range.

{\clubpenalty=10000
In the primary dataset, amplitude- and power-related descriptors from both muscles are selected less frequently as MVC increases, with none retained at the higher intensities. MI selection concerns discrimination between fatigue labels within a given MVC, rather than signal magnitude alone. With a fixed top-40 budget, reduced inclusion can also reflect competition from other descriptors. At higher MVC, additional motor-unit recruitment during fatigue may be more limited~\cite{fallentin_motor_1993}. This could reduce recruitment-driven increases in amplitude- and power-related descriptors and weaken the separation between fatigue states. Together, the stable core and the intensity-dependent selection patterns support evaluating physiologically motivated representations across multiple contraction intensities. Detailed selection statistics are provided in Supplementary Sections~S7 and S9.\par}

\subsection{Ablation studies}
Starting from full FatigueFormer, we separately remove the static branch, remove the temporal branch, replace cross-attention with self-attention, or omit the auxiliary classification losses $\mathcal{L}_{T}$ and $\mathcal{L}_{S}$ (multi-scale supervision). Figure~\ref{fig:ablation} summarizes the effects across protocols and MVC levels in both datasets. Significant changes are marked after correction for multiple comparisons~($q<0.05$).

\paragraph{Static branch}
Removing the static branch lowers performance at 80\% MVC in the primary dataset under both protocols, with a significant decrease under LOSO. By contrast, removal significantly improves within-subject performance at 20--60\% MVC in the primary dataset and at both intensities in SWIFTIES. These results suggest that within-window descriptor interactions complement temporal context in the high-intensity LOSO setting. In some within-subject settings, however, the additional interactions may fit unstable descriptor relationships and increase overfitting risk.

\paragraph{Temporal branch}
Removing the temporal branch lowers performance at every evaluated intensity under both protocols in both datasets, with most differences reaching statistical significance. This supports a consistent contribution from inter-window context. Under the within-subject protocol, the largest decrease occurs at the lowest evaluated MVC in each dataset. One possible explanation is that single-window fatigue cues are weaker at low intensities, making neighboring context more important for distinguishing fatigue states.

\paragraph{Cross-attention}
At 80\% MVC under LOSO in the primary dataset, replacing cross-attention with self-attention reduces macro-F1 by 9.26 percentage points, the numerically largest decrease among the four ablations in this condition. A significant decrease after replacement is observed only in this condition. Cross-trend exchange may therefore help combine complementary INC and DEC cues that independent within-stream modeling does not fully capture in this high-intensity cross-subject setting.

\paragraph{Multi-scale supervision}
Removing the auxiliary classification losses lowers performance in most conditions. The decreases are statistically significant at every evaluated intensity under within-subject evaluation in both datasets, whereas under LOSO, significance is observed only at 80\% MVC in the primary dataset. These results suggest that adding supervision to the temporal and static branches can further improve fatigue recognition. The gains are more consistent under the within-subject protocol. The auxiliary heads provide a direct classification objective for each branch, which may help preserve useful information in both representations when their contributions to the fused prediction are unequal.

\section{Conclusion and Future Work}
We presented FatigueFormer for sEMG-based fatigue recognition, combining physiologically motivated trend grouping with inter-window context and within-window feature interactions. Experiments on two cohorts show competitive performance across contraction intensities and evaluation protocols. A stable descriptor core recurs across cohorts, while feature selection varies with contraction intensity and component contributions depend on both intensity and evaluation protocol. Fine-tuning with more labeled target-user data also yields higher scores on complete target recordings, which include calibration samples.

Future work will examine whether fatigue estimates can help sEMG-based typing and gesture-recognition systems adapt during prolonged use. Joint modeling of intended actions and fatigue may help distinguish task-related signal changes from changes in the user's physiological state. We will also explore online recognition using only past and current data, together with adaptation that requires fewer target-user labels. Evaluation across recording sessions and during dynamic movements will assess robustness under more realistic interaction conditions.

\section*{Ethical Impact Statement}
The primary study was approved by the Medical Ethics Committee of the Department of Psychology and Behavioral Sciences, Zhejiang University (approval no.~[2025]~053). All participants provided written informed consent before acquisition, including consent for de-identified publication. SWIFTIES is used as an independent research cohort. The study evaluates RPE-derived exertion states for research use. Applications involving health or work-capacity decisions require dedicated validation. Subgroup analyses are exploratory given the cohort sizes and available demographic information.

\label{lastcontent}
\bibliographystyle{ieee}
\bibliography{references}

@inproceedings{10211798,
  author = {Yang, Rongkai and Deng, Hengzhang and Xu, Wenteng and Wang, Xinyi and Li, Chunmao},
  title = {Multi-stream {CNN-SVM} Hybrid Model for Gesture Recognition based on {sEMG} Signals},
  booktitle = {2023 6th International Conference on Electronics Technology (ICET)},
  year = {2023},
  pages = {1435--1440},
  publisher = {IEEE},
  doi = {10.1109/icet58434.2023.10211798}
}

@article{10285513,
  author = {Zabihi, Soheil and Rahimian, Elahe and Asif, Amir and Mohammadi, Arash},
  title = {{TraHGR}: Transformer for Hand Gesture Recognition via Electromyography},
  journal = {IEEE Transactions on Neural Systems and Rehabilitation Engineering},
  year = {2023},
  volume = {31},
  pages = {4211--4224},
  doi = {10.1109/tnsre.2023.3324252}
}

@book{Borg1998,
  author = {Borg, Gunnar},
  title = {Borg's Perceived Exertion and Pain Scales},
  publisher = {Human Kinetics},
  address = {Champaign, IL},
  year = {1998},
  isbn = {9780880116237}
}

@article{DANG2023112,
  author = {Dang, Yukun and Liu, Zitong and Yang, Xixin and Ge, Linqiang and Miao, Sheng},
  title = {A fatigue assessment method based on attention mechanism and surface electromyography},
  journal = {Internet of Things and Cyber-Physical Systems},
  year = {2023},
  volume = {3},
  pages = {112--120},
  doi = {10.1016/j.iotcps.2023.03.002}
}

@article{DeLuca1997,
  author = {De Luca, Carlo J.},
  title = {The Use of Surface Electromyography in Biomechanics},
  journal = {Journal of Applied Biomechanics},
  year = {1997},
  volume = {13},
  number = {2},
  pages = {135--163},
  doi = {10.1123/jab.13.2.135}
}

@book{Electromyography,
  editor = {Merletti, Roberto and Parker, Philip},
  title = {Electromyography: Physiology, Engineering, and Noninvasive Applications},
  year = {2004},
  publisher = {John Wiley \& Sons},
  doi = {10.1002/0471678384},
  isbn = {9780471675808}
}

@article{Musclefatigue,
  author = {Enoka, Roger M. and Duchateau, Jacques},
  title = {Muscle fatigue: what, why and how it influences muscle function},
  journal = {The Journal of Physiology},
  year = {2008},
  volume = {586},
  number = {1},
  pages = {11--23},
  doi = {10.1113/jphysiol.2007.139477}
}

@article{PHINYOMARK20127420,
  author = {Phinyomark, Angkoon and Phukpattaranont, Pornchai and Limsakul, Chusak},
  title = {Feature reduction and selection for {EMG} signal classification},
  journal = {Expert Systems with Applications},
  year = {2012},
  volume = {39},
  number = {8},
  pages = {7420--7431},
  doi = {10.1016/j.eswa.2012.01.102}
}

@inproceedings{Song_2019,
  author = {Song, Weiping and Shi, Chence and Xiao, Zhiping and Duan, Zhijian and Xu, Yewen and Zhang, Ming and Tang, Jian},
  title = {{AutoInt}: Automatic Feature Interaction Learning via Self-Attentive Neural Networks},
  booktitle = {Proceedings of the 28th ACM International Conference on Information and Knowledge Management},
  year = {2019},
  pages = {1161--1170},
  publisher = {ACM},
  doi = {10.1145/3357384.3357925}
}

@inproceedings{Wang_2020,
  author = {Wang, Lichen and Sun, Bin and Robinson, Joseph and Jing, Taotao and Fu, Yun},
  title = {{EV-Action}: Electromyography-Vision Multi-Modal Action Dataset},
  booktitle = {2020 15th IEEE International Conference on Automatic Face and Gesture Recognition (FG 2020)},
  year = {2020},
  pages = {160--167},
  publisher = {IEEE},
  doi = {10.1109/fg47880.2020.00018}
}

@inproceedings{Yu2024,
  author = {Yu, Jiahui and Yang, Jun and Zhu, Chang and Meng, Wei and Liu, Quan and Zhou, Zude},
  title = {Hybrid {CNN-LSTM-Transformer} Model for Robust Muscle Fatigue Detection during Rehabilitation Using {sEMG} Signals},
  booktitle = {2024 International Conference on Advanced Robotics and Mechatronics (ICARM)},
  year = {2024},
  pages = {51--56},
  publisher = {IEEE},
  doi = {10.1109/icarm62033.2024.10715764}
}

@article{article8,
  author = {Toledo-P{\'e}rez, Diana C. and Rodr{\'i}guez-Res{\'e}ndiz, Juvenal and G{\'o}mez-Loenzo, Roberto A. and Jauregui-Correa, J. C.},
  title = {Support Vector Machine-Based {EMG} Signal Classification Techniques: A Review},
  journal = {Applied Sciences},
  year = {2019},
  volume = {9},
  number = {20},
  pages = {4402},
  doi = {10.3390/app9204402}
}

@article{fallentin_motor_1993,
  author = {Fallentin, Nils and J{\o}rgensen, Kurt and Simonsen, Erik B.},
  title = {Motor unit recruitment during prolonged isometric contractions},
  journal = {European Journal of Applied Physiology and Occupational Physiology},
  year = {1993},
  volume = {67},
  number = {4},
  pages = {335--341},
  doi = {10.1007/bf00357632}
}

@inproceedings{gorishniy2023revisitingdeeplearningmodels,
  author = {Gorishniy, Yury and Rubachev, Ivan and Khrulkov, Valentin and Babenko, Artem},
  title = {Revisiting Deep Learning Models for Tabular Data},
  booktitle = {Advances in Neural Information Processing Systems},
  year = {2021},
  volume = {34},
  pages = {18932--18943},
  publisher = {Curran Associates, Inc.},
  url = {https://proceedings.neurips.cc/paper/2021/hash/9d86d83f925f2149e9edb0ac3b49229c-Abstract.html}
}

@inproceedings{kingma2017adammethodstochasticoptimization,
  author = {Kingma, Diederik P. and Ba, Jimmy},
  title = {{Adam}: A Method for Stochastic Optimization},
  booktitle = {International Conference on Learning Representations},
  year = {2015},
  url = {https://arxiv.org/abs/1412.6980}
}

@inproceedings{vaswani2023attentionneed,
  author = {Vaswani, Ashish and Shazeer, Noam and Parmar, Niki and Uszkoreit, Jakob and Jones, Llion and Gomez, Aidan N. and Kaiser, {\L}ukasz and Polosukhin, Illia},
  title = {Attention Is All You Need},
  booktitle = {Advances in Neural Information Processing Systems},
  year = {2017},
  volume = {30},
  pages = {5998--6008},
  publisher = {Curran Associates, Inc.},
  url = {https://proceedings.neurips.cc/paper/2017/hash/3f5ee243547dee91fbd053c1c4a845aa-Abstract.html}
}

@article{Ming2014FatigueTrends,
  author = {Ming, Dong and Wang, Xin and Xu, Rui and Qiu, Shuang and Zhao, Xin and Qi, Hongzhi and Zhou, Peng and Zhang, Lixin and Wan, Baikun},
  title = {{sEMG} feature analysis on forearm muscle fatigue during isometric contractions},
  journal = {Transactions of Tianjin University},
  year = {2014},
  volume = {20},
  number = {2},
  pages = {139--143},
  doi = {10.1007/s12209-014-2181-2}
}

@article{Arjunan2014FatigueFeatures,
  author = {Arjunan, Sridhar P. and Kumar, Dinesh K. and Naik, Ganesh},
  title = {Computation and Evaluation of Features of Surface Electromyogram to Identify the Force of Muscle Contraction and Muscle Fatigue},
  journal = {BioMed Research International},
  year = {2014},
  volume = {2014},
  pages = {1--6},
  doi = {10.1155/2014/197960}
}

@article{Eddy2024,
  author = {Eddy, Ethan and Campbell, Evan and Bateman, Scott and Scheme, Erik},
  title = {Big data in myoelectric control: large multi-user models enable robust zero-shot {EMG}-based discrete gesture recognition},
  journal = {Frontiers in Bioengineering and Biotechnology},
  year = {2024},
  volume = {12},
  pages = {1463377},
  doi = {10.3389/fbioe.2024.1463377}
}

@article{Cerqueira2024,
  author = {Cerqueira, Sara M. and Vilas Boas, Rita and Figueiredo, Joana and Santos, Cristina P.},
  title = {A Comprehensive Dataset of Surface Electromyography and Self-Perceived Fatigue Levels for Muscle Fatigue Analysis},
  journal = {Sensors},
  year = {2024},
  volume = {24},
  number = {24},
  pages = {8081},
  doi = {10.3390/s24248081}
}

@article{Levenez2008Coactivation,
  author = {L{\'e}v{\'e}nez, Morgan and Garland, S. Jayne and Klass, Malgorzata and Duchateau, Jacques},
  title = {Cortical and spinal modulation of antagonist coactivation during a submaximal fatiguing contraction in humans},
  journal = {Journal of Neurophysiology},
  year = {2008},
  volume = {99},
  number = {2},
  pages = {554--563},
  doi = {10.1152/jn.00963.2007}
}

@article{article3,
  author = {Cifrek, Mario and Medved, Vladimir and Tonkovi{\'c}, Stanko and Ostoji{\'c}, Sa{\v{s}}a},
  title = {Surface {EMG} based muscle fatigue evaluation in biomechanics},
  journal = {Clinical Biomechanics},
  year = {2009},
  volume = {24},
  number = {4},
  pages = {327--340},
  doi = {10.1016/j.clinbiomech.2009.01.010}
}

@article{KARTHICK201642,
  author = {Karthick, P. A. and Ramakrishnan, S.},
  title = {Surface electromyography based muscle fatigue progression analysis using modified {B} distribution time--frequency features},
  journal = {Biomedical Signal Processing and Control},
  year = {2016},
  volume = {26},
  pages = {42--51},
  doi = {10.1016/j.bspc.2015.12.007}
}

@article{article7,
  author = {Karthick, P. A. and Ghosh, Diptasree Maitra and Ramakrishnan, S.},
  title = {Surface electromyography based muscle fatigue detection using high-resolution time-frequency methods and machine learning algorithms},
  journal = {Computer Methods and Programs in Biomedicine},
  year = {2018},
  volume = {154},
  pages = {45--56},
  doi = {10.1016/j.cmpb.2017.10.024}
}

@article{Luttmann2000JASA,
  author = {Luttmann, Alwin and J{\"a}ger, Matthias and Laurig, Wolfgang},
  title = {Electromyographical indication of muscular fatigue in occupational field studies},
  journal = {International Journal of Industrial Ergonomics},
  year = {2000},
  volume = {25},
  number = {6},
  pages = {645--660},
  doi = {10.1016/S0169-8141(99)00053-0}
}

@article{Zhang2021MFFNet,
  author = {Zhang, Yongqing and Chen, Siyu and Cao, Wenpeng and Guo, Peng and Gao, Dongrui and Wang, Manqing and Zhou, Jiliu and Wang, Ting},
  title = {{MFFNet}: Multi-dimensional Feature Fusion Network based on attention mechanism for {sEMG} analysis to detect muscle fatigue},
  journal = {Expert Systems with Applications},
  year = {2021},
  volume = {185},
  pages = {115639},
  doi = {10.1016/j.eswa.2021.115639}
}

@article{Yin2026WristFatigue,
  author = {Yin, Kun and Lopez, Guillaume},
  title = {Deep-learning Based Wrist Fatigue Detection Using Multi-Domain {sEMG} Features},
  journal = {International Journal of Activity and Behavior Computing},
  year = {2026},
  volume = {2026},
  number = {1},
  pages = {1--18},
  doi = {10.60401/ijabc.145}
}

@article{Wang2021LSTMFatigue,
  author = {Wang, Junhong and Sun, Shaoming and Sun, Yining},
  title = {A Muscle Fatigue Classification Model Based on {LSTM} and Improved Wavelet Packet Threshold},
  journal = {Sensors},
  year = {2021},
  volume = {21},
  number = {19},
  pages = {6369},
  doi = {10.3390/s21196369}
}

@article{Zhang2024,
  author = {Zhang, Guofu and Yang, Banghua and Zan, Peng and Zhang, Dingguo},
  title = {Multilevel Assessment of Exercise Fatigue Utilizing Multiple Attention and Convolution Network ({MACNet}) Based on Surface Electromyography},
  journal = {IEEE Transactions on Neural Systems and Rehabilitation Engineering},
  year = {2025},
  volume = {33},
  pages = {243--254},
  doi = {10.1109/tnsre.2024.3523332}
}

@misc{OSullivan2024SWIFTIES,
  author = {O'Sullivan, Patricia},
  title = {{SWIFTIES} (Subject and Wearables data for Investigation of Fatiguing Tasks In Extension/flexion of Shoulder/elbow)},
  howpublished = {Zenodo},
  year = {2024},
  month = {September},
  note = {Dataset, version v5, released September 17, 2024. DOI: \url{https://doi.org/10.5281/zenodo.15172815}},
  doi = {10.5281/zenodo.15172815},
  url = {https://zenodo.org/records/15172815}
}

@article{Kaifosh2025GenericNeuromotor,
  author = {Kaifosh, Patrick and Reardon, Thomas R. and {CTRL-labs at Reality Labs}},
  title = {A generic non-invasive neuromotor interface for human-computer interaction},
  journal = {Nature},
  year = {2025},
  volume = {645},
  number = {8081},
  pages = {702--711},
  doi = {10.1038/s41586-025-09255-w},
  url = {https://doi.org/10.1038/s41586-025-09255-w}
}

@inproceedings{Salter2024Emg2pose,
  author = {Salter, Sasha and Warren, Richard and Schlager, Collin and Spurr, Adrian and Han, Shangchen and Bhasin, Rohin and Cai, Yujun and Walkington, Peter and Bolarinwa, Anuoluwapo and Wang, Robert and Danielson, Nathan and Merel, Josh and Pnevmatikakis, Eftychios and Marshall, Jesse},
  title = {{emg2pose}: A Large and Diverse Benchmark for Surface Electromyographic Hand Pose Estimation},
  booktitle = {Advances in Neural Information Processing Systems},
  volume = {37},
  pages = {55703--55728},
  publisher = {Curran Associates, Inc.},
  year = {2024},
  doi = {10.52202/079017-1770},
  url = {https://proceedings.neurips.cc/paper_files/paper/2024/file/64f8884f6ba3d9ace5bb647c4f917896-Paper-Datasets_and_Benchmarks_Track.pdf}
}

@inproceedings{Sivakumar2024Emg2qwerty,
  author = {Sivakumar, Viswanath and Seely, Jeffrey and Du, Alan and Bittner, Sean R and Berenzweig, Adam and Bolarinwa, Anuoluwapo and Gramfort, Alexandre and Mandel, Michael I},
  title = {{emg2qwerty}: A Large Dataset with Baselines for Touch Typing using Surface Electromyography},
  booktitle = {Advances in Neural Information Processing Systems},
  volume = {37},
  pages = {91373--91389},
  publisher = {Curran Associates, Inc.},
  year = {2024},
  doi = {10.52202/079017-2899},
  url = {https://proceedings.neurips.cc/paper_files/paper/2024/file/a64d53074d011e49af1dfc72c332fe4b-Paper-Datasets_and_Benchmarks_Track.pdf}
}
\end{document}


\onecolumn
\pagestyle{plain}
\maketitle
\thispagestyle{plain}
This supplement documents the methods and supporting analyses for fatigue-state recognition across participants and contraction intensities.
\setcounter{figure}{0}
\setcounter{table}{0}

\noindent\textbf{Reading guide.} Sections~S1--S2 describe acquisition and descriptors; Section~S3 specifies model inputs, training and evaluation, including the complete-TEST adaptation results. Sections~S4--S8 provide participant-level and cross-cohort analyses and detailed statistics, followed by attention and feature-selection maps in Section~S9. Main-paper references are identified explicitly.

\section*{S1. Data Acquisition and Experimental Protocol}

The primary cohort comprised 30 healthy adults (15 male and 15 female). This section expands the acquisition protocol in main-paper Section~\MainDatasetSection{} and gives the class-specific window counts.

\paragraph*{Experimental setup}
Participants were seated with the upper arm stabilized and performed sustained isometric elbow flexion using the force-feedback apparatus shown in main-paper Figure~\MainAcquisitionFigure{}. Maximum voluntary contraction (MVC) was measured before the fatigue trials to set individual target forces at 20\%, 40\%, 60\% and 80\% MVC. Participants maintained each target using real-time visual feedback until they could no longer sustain at least 90\% of the prescribed force.

After skin cleaning with alcohol, bipolar electrodes were placed over the biceps brachii and triceps brachii muscle bellies and secured with gauze. Both channels were recorded at 2,000\,Hz from contraction onset to task failure, then band-pass filtered at 20--450\,Hz and notch filtered at 50\,Hz.

\paragraph*{Subjective fatigue labeling}
Participants reported perceived exertion in real time using Borg's RPE scale (6--20). The device timestamped each report for alignment with the sEMG recordings. Reports were mapped to the three fatigue states using the thresholds in main-paper Section~\MainMethodSection{}.

\paragraph*{Ethical compliance}
All participants provided written informed consent before data collection, including permission for the acquisition photograph. The ethics-board decision and approval identifier are reported in the main paper's Ethical Impact Statement.

\begin{table}[!htbp]
\centering
\caption{Pooled window counts by fatigue state and MVC level in the primary cohort, before the exclusions required for valid within-subject folds.}
\label{tab:sample_distribution}
\begin{tabular*}{0.72\textwidth}{@{\extracolsep{\fill}}lcccc@{}}
\toprule
MVC Level & Relaxed & Exerted & Fatigued & Total \\
\midrule
20\% & 6313 & 9867 & 16417 & 32597 \\
40\% & 1475 & 3331 & 4318 & 9124 \\
60\% & 775  & 1408 & 1955 & 4138 \\
80\% & 402  & 813  & 1603 & 2818 \\
\bottomrule
\end{tabular*}
\end{table}

Window counts decrease with MVC (Supplementary Table~\ref{tab:sample_distribution}), reflecting shorter sustainable recording durations at higher intensities. These pooled counts describe data availability before protocol-specific exclusions; effective within-subject participant counts are reported in Sections~S5 and S8.

\section*{S2. Feature Definitions and Trend Analysis}
Supplementary Table~\ref{tab:feature_definitions} defines the 31 descriptors computed per muscle channel, giving 62 inputs across the biceps brachii (BB) and triceps brachii (TB). The INC/DEC assignments and descriptor families below support the feature analysis in main-paper Section~\MainResultsSection{}.

In the table, $x_i$ denotes a preprocessed sample in a window of $N$ samples acquired at $f_s$, with mean $\bar{x}$ and standard deviation $\sigma$. The pair $\{f_k,P_k\}$ denotes the one-sided periodogram; $X=\mathcal{F}x$ is the discrete Fourier transform, $\mathcal{S}_{k,m}$ the STFT spectrogram, and $D_j$ the level-$j$ wavelet detail coefficients. The operators $\Delta$ and $\mathbb{1}[\cdot]$ denote the first difference and indicator function. \texttt{MF} and \texttt{MDF} both measure median frequency, using linear interpolation and the nearest cumulative-power bin, respectively. \texttt{IMPF} uses the \texttt{IMNF} definition because mean power frequency and mean frequency are synonymous.

\begin{table}[!htbp]
\small
\centering
\caption{Definitions and trend assignments of the 31 sEMG descriptors computed per muscle channel (INC, increasing; DEC, decreasing).}
\label{tab:feature_definitions}
\setlength{\tabcolsep}{4pt}
\begin{tabular}{@{}llp{0.67\textwidth}@{}}
\toprule
Descriptor & Trend & Definition \\
\midrule
\texttt{iEMG}     & INC & $\dfrac{f_s}{N}\sum_i |x_i|$ \\
\texttt{RMS}      & INC & $\sqrt{\tfrac{1}{N}\sum_i x_i^2}$ \\
\texttt{MAV}      & INC & $\tfrac{1}{N}\sum_i |x_i|$ \\
\texttt{MCV}      & INC & $\tfrac{1}{N}\sum_i x_i^2$ \ ($=\mathrm{RMS}^2$) \\
\texttt{DASDV}    & INC & $\sqrt{\tfrac{1}{N-1}\sum_i (x_{i+1}-x_i)^2}$ \\
\texttt{SMR}      & INC & $\big(\sum_k f_k^{-1}P_k\big)\big/\big(\sum_k f_k^{5}P_k\big)$, \ $f_k\in[f_{5\%},f_{95\%}]$ \\
\texttt{FSM2}     & INC & $\sum_k f_k^{2}P_k$ \\
\texttt{TP}       & INC & $\sum_k P_k$ \\
\texttt{WIRM1551} & INC & $\int |\mathcal{F}(D_5)|^2 f^{-1}\,df \big/ \int |\mathcal{F}(D_1)|^2 f^{5}\,df$ \ (sym5; $f\!\in\![10,500]$\,Hz) \\
\texttt{WIRM1522} & INC & $\int |\mathcal{F}(D_5)|^2 df \big/ \int |\mathcal{F}(D_2)|^2 f^{2}\,df$ \ (db5; $f\!\in\![10,500]$\,Hz) \\
\texttt{WIRE51}   & INC & $\sum (D_5)^2 \big/ \sum (D_1)^2$ \ (sym5) \\
\texttt{WIRW51}   & INC & $\sum (\Delta D_5)^2 \big/ \sum (\Delta D_1)^2$ \ (sym5) \\
\texttt{ERHL}     & INC & $\big(\sum |D_5|^2 + \sum |D_4|^2\big)\big/\sum |D_1|^2$ \ (db5) \\
\texttt{DET}      & INC & fraction of recurrence points on diagonal lines of length $\ge\ell_{\min}$ ($m{=}2$, $\tau{=}17$, $\varepsilon{=}0.05$) \\
\texttt{ACC}      & INC & $\sum_i (x_i-\bar x)(x_{i+1}-\bar x)\big/\sum_i (x_i-\bar x)^2$ \\
\midrule
\texttt{ZC}       & DEC & $\sum_i \mathbb{1}[\operatorname{sgn}(x_i)\ne\operatorname{sgn}(x_{i+1})]$ \\
\texttt{WA}       & DEC & $\sum_i \mathbb{1}[\,|x_{i+1}-x_i|>\sigma\,]$ \\
\texttt{SSC}      & DEC & $\sum_i \mathbb{1}[(x_i-x_{i-1})(x_i-x_{i+1})>0]$ \\
\texttt{MPF}      & DEC & $\big(\sum_k f_k P_k\big)\big/\big(\sum_k P_k\big)$ \ (mean frequency) \\
\texttt{MDF}      & DEC & $f_{\mathrm{med}}$ s.t.\ $\sum_{f_k\le f_{\mathrm{med}}}P_k=\tfrac12\sum_k P_k$ \ (median frequency) \\
\texttt{MF}       & DEC & median frequency (half-power point by linear interpolation) \\
\texttt{IMNF}     & DEC & $\tfrac{1}{M}\sum_m \big(\sum_k f_k \mathcal{S}_{k,m}\big/\sum_k \mathcal{S}_{k,m}\big)$, \ $f_k\in[10,300]$\,Hz \\
\texttt{IMPF}     & DEC & instantaneous mean-power frequency (STFT) \\
\texttt{IMF}      & DEC & $\tfrac{1}{M}\sum_m f_{\mathrm{med}}(t_m)$ \ (STFT per-frame median frequency) \\
\texttt{WENT}     & DEC & $\tfrac{1}{N-1}\sum_i \mathbb{1}[\,|x_{i+1}-x_i|>\sigma\,]$ \\
\texttt{FD}       & DEC & Higuchi fractal dimension ($k_{\max}=15$) \\
\texttt{AE}       & DEC & approximate entropy (ApEn) \\
\texttt{SE}       & DEC & sample entropy (SampEn) \\
\texttt{LZC}      & DEC & $c(N)\log_2 N / N$, \ $c(N)=$ LZ76 distinct-substring count \\
\texttt{BE}       & DEC & $-\sum_k p_k\ln p_k$, \ $p_k=|X_k|^2/\sum_k |X_k|^2$ \ (spectral entropy) \\
\texttt{BSE}      & DEC & $-\sum_{b=1}^{5} q_b\ln q_b$, \ $q_b=E_b/\sum_b E_b$ \ ($E_b=$ energy in 5 equal sub-bands) \\
\bottomrule
\end{tabular}

\end{table}

\paragraph*{Trend-based assignment}
As described in main-paper Section~\MainMethodSection{}, candidate descriptors are regressed against ordinal fatigue labels (relaxed = 0, exerted = 1, fatigued = 2), retaining 31 with significant slopes ($p<0.05$). Positive and negative slopes define the 15 INC and 16 DEC descriptors, respectively. This analysis pools labeled observations across participants and MVC levels (20--80\%) to establish a common assignment for both muscle channels. The descriptor set and INC/DEC lists remain fixed across folds; full FatigueFormer's subsequent MI selection is fitted separately within each training fold.

\paragraph*{Descriptor families}
The descriptors are grouped into 13 families by signal property or algorithmic construction. The six \textbf{INC families} are amplitude-related measures (\texttt{iEMG, RMS, MAV, MCV, DASDV}), the spectral-moment ratio (\texttt{SMR}), frequency-domain power (\texttt{FSM2, TP}), wavelet-coefficient ratios (\texttt{WIRM1551, WIRM1522, WIRE51, WIRW51}), the wavelet high/low band-energy ratio (\texttt{ERHL}), and recurrence/autocorrelation (\texttt{DET, ACC}). The amplitude/power analysis in the main text and Section~S7 uses these five amplitude measures plus \texttt{FSM2} and \texttt{TP}.

The seven \textbf{DEC families} are zero-crossing rate (\texttt{ZC}), waveform-complexity counts (\texttt{WA, SSC}), the threshold-crossing proportion (\texttt{WENT}), mean/median frequency (\texttt{MDF, MPF, MF}), instantaneous frequency measures (\texttt{IMF, IMNF, IMPF}), fractal dimension (\texttt{FD}), and complexity/spectral entropies (\texttt{AE, SE, LZC, BE, BSE}). These family labels describe signal properties rather than uniquely identifying physiological mechanisms. Their interpretation depends on the contraction context, since fatigue involves interacting neural and muscular processes~\cite{Musclefatigue}.

\paragraph*{Representative temporal trends}

\begin{figure}[!htbp]
\centering
\includegraphics[width=\linewidth]{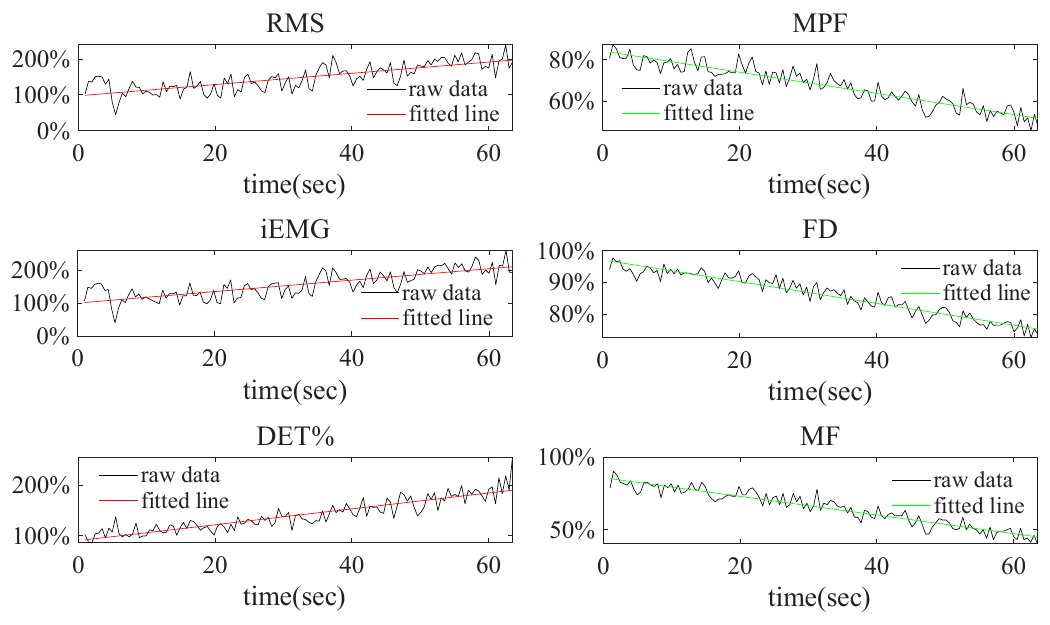}
\caption{Temporal trajectories of six descriptors from Supplementary Table~\ref{tab:feature_definitions}. Black curves show subject-averaged trajectories and colored lines show linear fits. RMS, iEMG and DET illustrate INC trends; MPF, FD and MF illustrate DEC trends.}
\label{fig:feature_trends}
\end{figure}

Supplementary Fig.~\ref{fig:feature_trends} illustrates temporal trend directions during the task. The INC/DEC assignment itself uses the ordinal-label regression described above; these example trajectories do not replace that statistical criterion.

\FloatBarrier

\section*{S3. Model and baseline implementation details}

This section details the benchmark inputs and training settings, the Feature-LT and Full--Swap50 controls, and the target-participant adaptation protocol. Statistical conventions are given at the end of the section.

\paragraph*{FatigueFormer implementation}
MI selection uses the three-nearest-neighbor estimator in \texttt{mutual\_info\_classif}, with continuous descriptors and discrete labels. Selected indices are restored to their original order before INC/DEC partitioning. Feature-tokenizer parameters are initialized from $\mathcal{N}(0,1)$. Prototype means remain in the gradient graph, with each embedding contributing to its own class mean. The prototype denominator includes only classes present in the mini-batch; batches with fewer than two classes contribute zero prototype loss. Validation and test predictions use the joint head.

\paragraph*{Primary-cohort benchmark inputs and training}
The neural baselines in main-paper Section~\MainEvaluationSection{} are implemented in PyTorch and trained from scratch. AutoInt and RBF-SVM use the same 62-descriptor representation supplied to FatigueFormer before MI selection. CNN--SVM, CLT and MACNet use dual-channel 2-s raw sEMG segments of shape $4000\!\times\!2$. MACNet divides each segment into 16 overlapping 1,000-sample subwindows (stride 200), each inheriting the segment label. Raw inputs receive no feature-level z-scoring; architecture-internal normalization layers are retained. Corresponding backbone or hidden dimensions are set to $D\!=\!256$.

The original primary-cohort neural benchmarks run on an NVIDIA RTX 4070 using Adam~\cite{kingma2017adammethodstochasticoptimization}, learning rate $10^{-4}$, weight decay $10^{-3}$ and batch size 32. Early stopping uses validation accuracy, measured at the joint head for FatigueFormer. Feature-LT and adaptation settings are specified separately below.

Both FatigueFormer variants use the same backbone and three classification heads. Training allows 200 epochs with patience 30 under LOSO and 120 epochs with patience 20 within subject. LOSO uses participant-wise full-record z-scoring; the full variant also fits a training scaler after MI selection. Within subject, both variants fit a scaler on the current training fold and use seeds 0, 1 and 2. Up to five stratified folds are formed according to class support, and splits missing a training class are skipped. Participant scores average the available fold--seed cells. Within-subject results for the original five external comparators aggregate seeds 0--4.

\paragraph*{Primary-cohort RBF-SVM}
Each LOSO fold fits StandardScaler and the SVM on all 29 non-test participants, without a validation holdout. FatigueFormer and CNN--SVM instead reserve one source participant for validation and train on 28. RBF-SVM uses the RBF kernel, fixed inherited $C=10$, \texttt{gamma='scale'} and \texttt{class\_weight='balanced'}, with other parameters at the scikit-learn defaults. Gamma is calculated from the variance of standardized source-training data. No grid search, validation-based hyperparameter selection or nested cross-validation is used.

\paragraph*{SWIFTIES benchmark inputs and aggregation}
The seven-model benchmark in main-paper Table~\MainPerformanceTable{} uses 32 participants at 25\% and 45\% MVC and seed 0. Full-model scores reuse the existing runs; base and the five external comparators use aligned outer test targets. Within-subject scores average five folds per participant before averaging participants equally; LOSO averages the 32 held-out participants.

FatigueFormer uses five-window descriptor sequences, whereas AutoInt and RBF-SVM use the target window's 62 descriptors. CLT, CNN--SVM and MACNet receive contiguous two-channel inputs of 1,536 samples, reconstructed from the five overlapping windows without duplicating overlap. MACNet uses 16 subwindows of 396 samples (stride 76) and majority voting for the target prediction. Only GPU-backend CNN--SVM runs enter the reported means; legacy CPU runs are excluded. Raw inputs receive no descriptor-wise z-scoring, and CNN--SVM fits a scaler on training encoder features.

\paragraph*{SWIFTIES normalization, splits and training}
Feature models in this seven-model benchmark inherit full-record participant z-scoring. Base additionally fits a training-fold scaler within subject, but not under LOSO; full uses training-fold scaling under both protocols. Neural LOSO models use 30 training participants, one validation participant and one test participant; RBF-SVM trains on the other 31. Within-subject random five-fold splits are applied after constructing overlapping windows. The inner training/validation split is 80/20 for base/full and 85/15 for the five external comparators; RBF-SVM does not fit on the reserved validation samples.

Epoch limits/patience for LOSO and within-subject training, respectively, are 100/15 and 100/15 for base/full; 300/50 and 80/12 for AutoInt; 100/15 and 40/8 for CLT; 80/15 and 40/8 for the CNN--SVM encoder; and 60/10 and 30/6 for MACNet. Batch sizes are 32, 32, 16, 32 and 8 in the same model order. Early stopping uses validation accuracy, measured at the subwindow level for MACNet. Thus the benchmark compares the reported model-specific configurations rather than architectures under a uniform training budget.

\paragraph*{Feature-LT controls}
Main-paper Table~III reports Feature-LT62: all 62 descriptors without MI selection, arranged as five consecutive windows in a $(B,5,62)$ input tensor. The third window supplies the target label. Feature-LT40 instead uses the full model's training-fold MI top-40 panel and $(B,5,40)$ inputs. In both controls, the sequence axis represents windows rather than descriptor indices. They reuse the frozen full-model training, validation and test partitions and fit StandardScaler only on the training fold.

The network comprises a single-layer bidirectional LSTM with 128 hidden units per direction, two Transformer encoder layers ($D=256$, four heads, feed-forward dimension 512, GELU, dropout 0.1), temporal mean pooling and a $256$-to-$3$ linear classifier. Only the LSTM input dimension differs: Feature-LT40 has 1,229,059 parameters and Feature-LT62 has 1,251,587. Neither uses a CNN, CLS token, additional positional encoding, INC/DEC streams, a static branch or prototype loss.

Both datasets use Adam with learning rate $10^{-4}$, zero weight decay and batch size 16. Training uses a single-head class-balanced cross-entropy loss, with weights estimated on the training set, and no label smoothing, learning-rate scheduling or additional gradient clipping. LOSO allows 100 epochs with patience 15; within-subject training allows 40 epochs with patience 8. A checkpoint is saved only when validation accuracy strictly improves, and testing uses that best checkpoint. Each version uses training seed 0 only and covers 12 dataset--protocol--MVC conditions with 1,016 valid folds under the frozen splits; these folds are not independent random repetitions. LOSO averages participants equally. Within subject, valid folds are first averaged within participant, followed by equal averaging across participants.

\paragraph*{Fixed partial trend exchange (Full--Swap50)}
Full--Swap50 exchanges part of the INC and DEC assignments after MI selects the same 40 descriptors as full FatigueFormer. It preserves the selected panel rather than performing Random-40 feature selection or a uniformly random repartition. For retained stream sizes $n_{\mathrm{inc}}$ and $n_{\mathrm{dec}}$, it samples without replacement and exchanges
\[
k=\operatorname{round}\!\left(0.5\min(n_{\mathrm{inc}},n_{\mathrm{dec}})\right)
\]
descriptors from each stream. Rounding follows Python's ties-to-even rule. NumPy \texttt{RandomState(0)} samples INC first and DEC second, after which each resulting stream is sorted by the original global feature indices. The original stream sizes are retained, without imposing a 20/20 split.

Each fold's mapping is frozen before training and shared across its training, validation and test data, both branches and subsequent adaptation. It is not redrawn by epoch or batch. Different folds may select different MI panels; identical panels produce identical mappings. The full architecture, three-head losses and 8,129,161 parameters are unchanged. Grouping seed 0 and training seed 0 form a single configuration, without repeated random groupings or outcome-based seed selection. The control therefore tests partial disruption of trend assignments, not removal of all trend information.

\paragraph*{Adaptation data and calibration budgets}
The adaptation models are full FatigueFormer, CNN--SVM and RBF-SVM. For a model's native target-participant sequence $T$, let $N=|T|$. Nominal 20\% and 40\% calibration sets initially contain the first $\lfloor N/5\rfloor$ and $\lfloor2N/5\rfloor$ samples, and the common evaluation subset $E$ starts at zero-based index $\lfloor2N/5\rfloor+5$. Budgets therefore use the original target sample count as their denominator.

Class-coverage adjustments add missing calibration classes using the earliest available consecutive same-class runs, trying lengths five down to one. If $E$ lacks a class, the procedure attempts to reserve the latest such run, again trying lengths five down to one, and removes its guard region without losing classes already represented in calibration. The final sets satisfy $\mathrm{cal20}\subseteq\mathrm{cal40}$, with calibration and $E$ separated by more than five local sample indices. Actual calibration fractions can increase or decrease, and the two sets coincide in a small number of folds. These adjustments can use labels outside the initial prefix. CSV order has not been verified as acquisition-time order, so the partitions are index-based rather than validated chronological splits.

Within each model, the 0\%, 20\% and 40\% stages use the same complete original TEST and the same frozen $E$. Main-paper Table~IV reports complete-TEST scores, which include calibration samples and do not measure independent target-user generalization. Scores on calibration-excluded $E$ are not reported in the current manuscript. Full FatigueFormer retains full-record participant normalization: 0\% means no target calibration labels, not absence of all target-data access.

\paragraph*{Model-specific adaptation samples}
In the primary dataset, full FatigueFormer uses MI-selected five-window sequences, CNN--SVM uses saved dual-channel 4,000-sample raw segments, and RBF-SVM uses single-window 62-dimensional vectors. Participants correspond across models, but sample counts, signal support and local indices differ. Each model constructs its own calibration sets and $E$, without reusing Full's local indices. Cross-model comparisons are consequently descriptive, not paired by sample. SWIFTIES outer target ordering can be aligned for Full and CNN--SVM; RBF-SVM adaptation was not run on that dataset.

\paragraph*{Neural adaptation settings}
For both full FatigueFormer and the CNN encoder, the 20\% and 40\% runs start independently from the same best source-validation checkpoint with new Adam optimizers. All parameters are updated using only the corresponding target calibration data, with learning rate $10^{-5}$, batch size 32 and exactly ten epochs. The final tenth-epoch weights are retained, without target-validation early stopping; the 40\% run does not continue the 20\% run.

\paragraph*{Full FatigueFormer adaptation}
Source LOSO training allows 200 epochs/patience 30 in the primary dataset and 100/15 in SWIFTIES, selecting the highest source-validation accuracy. Adaptation uses weight decay $10^{-3}$ and keeps the MI panel, fitted scaler and source class weights fixed. The objective sums the three class-balanced cross-entropy losses with label smoothing 0.1 and $0.5$ times the prototype loss with temperature 0.1.

\paragraph*{CNN--SVM adaptation}
Source encoder training allows 80 epochs with patience 15 and selects the highest source-validation accuracy. Adaptation uses zero weight decay and updates all CNN and auxiliary-head parameters; batch-normalization statistics are also updated. Each stage then re-extracts embeddings using its updated CNN. Its scaler and gamma are estimated only from source embeddings, and an RBF-SVM with $C=10$ is refitted on source plus the corresponding calibration embeddings, keeping source-global class weights fixed. SVM fitting parameters retain FP32 precision, with FP64 native/compact inference cross-checks.

\paragraph*{RBF-SVM adaptation}
The source model uses all 29 non-test primary participants. Adaptation refits RBF-SVM on source plus cal20 or cal40, without gradient updates or an epoch schedule. The 62-dimensional inputs, source-fitted scaler, $C=10$, gamma and source-global class weights stay fixed across the three stages. Gamma is explicitly computed as $1/[62\,\mathrm{Var}(X_{\mathrm{source},z})]$, matching the original \texttt{scale} rule. Fitting and inference both use GPU FP64, unlike CNN--SVM's FP32 fitting. Neural source-validation checkpoints do not select either SVM's $C$ or gamma.

\paragraph*{Statistical inference and reporting}
Benchmark macro-F1 is averaged within participant and then equally across participants. For primary-cohort external comparisons and component analyses, 95\% confidence intervals (CIs) use 10,000 participant-level BCa bootstrap resamples. Paired one-sided Wilcoxon tests include participants with valid scores for both methods at the same intensity, with Benjamini--Hochberg correction within each MVC. For these paired comparisons, $q$ is the adjusted value and $r$ the matched-pairs rank-biserial effect size. Component tests are conducted separately for each protocol and MVC; reverse-direction static-removal tests are identified explicitly. CIs describe participant variation, and overlapping LOSO training sets are not independent replications. The independent-group sex comparisons use the procedure stated in Section~S6.

Tables~\ref{tab:comparison}--\ref{tab:generalization} extend the primary-cohort comparisons in main-paper Table~\MainPerformanceTable{}. Tables~\ref{tab:ablation_mvc}--\ref{tab:ablation_loso} and Table~\ref{tab:sw_ablation} give the primary-cohort and SWIFTIES ablation statistics underlying main-paper Figure~\MainAblationFigure{}. Full/base, primary-cohort within-subject RBF-SVM, SWIFTIES external-model, Feature-LT62, Full--Swap50 and complete-TEST adaptation comparisons are descriptive, without paired-test statistics. In ablation tables, n.s.\ denotes $q\geq0.05$ when no numerical value is shown, and $\dagger$ marks a significant reverse-direction test. SWIFTIES ablation tests are reported separately in Section~S7.

\section*{S4. Per-class performance across MVC levels}

\begin{table}[!htbp]
\centering
\caption{Per-class F1 (\%) of full FatigueFormer in the primary cohort under within-subject cross-validation with up to five folds and three seeds. Macro-F1 is the mean of the three class-specific F1 values.}
\label{tab:per_class_f1_singlecol}
\begin{tabular*}{\textwidth}{@{\extracolsep{\fill}}lcccc@{}}
\toprule
\multirow{2}{*}{Class} & \multicolumn{4}{c}{MVC level (\%)}  \\
\cmidrule(lr){2-5}
 & 20 & 40 & 60 & 80 \\
\midrule
Relaxed  & 91.35 & 86.19 & 71.40 & 86.16 \\
Exerted  & 94.53 & 93.77 & 86.79 & 88.23 \\
Fatigued & 93.03 & 95.89 & 87.96 & 92.04 \\
\midrule
Macro-F1 (Table~\ref{tab:comparison}) & 92.97 & 91.95 & 82.05 & 88.81 \\
\bottomrule
\end{tabular*}
\end{table}

Supplementary Table~\ref{tab:per_class_f1_singlecol} separates full FatigueFormer's within-subject performance by RPE-derived state. The relaxed class has the lowest F1 at every MVC level, with the largest class difference at 60\% MVC. These descriptive results complement the model-level macro-F1 comparisons in main-paper Table~\MainPerformanceTable{} and the detailed statistics in Supplementary Table~\ref{tab:comparison}.

\section*{S5. Participant-level variability and outliers}

\begin{figure}[!htbp]
\centering
\includegraphics[width=\linewidth]{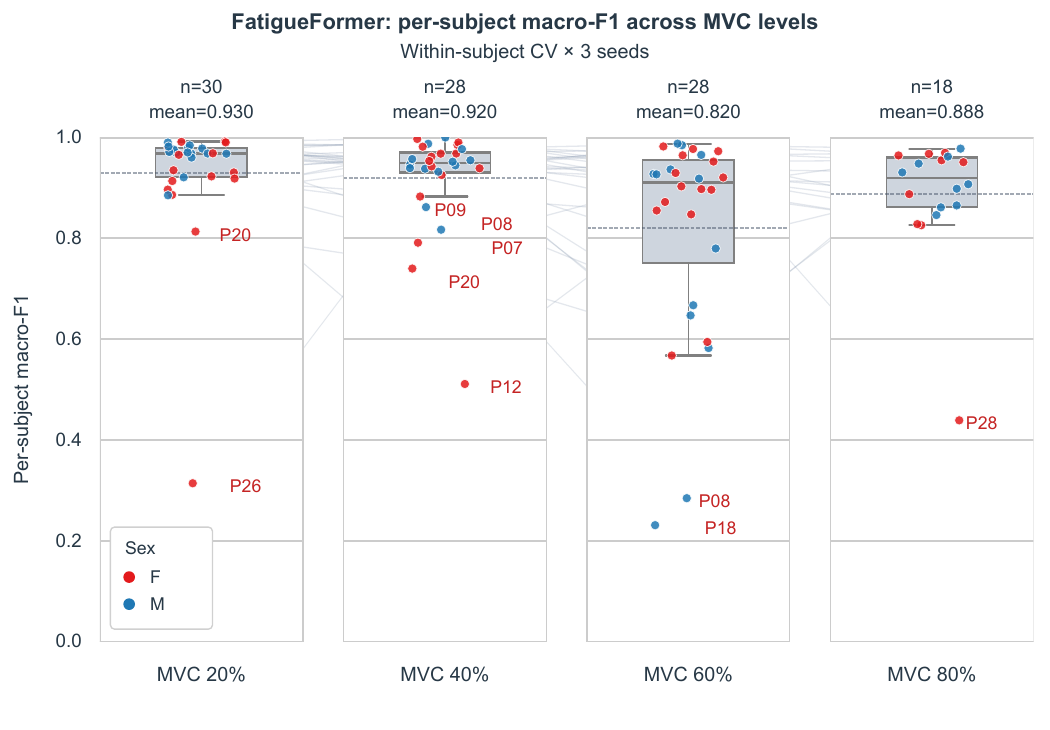}
\caption{Participant-level macro-F1 of full FatigueFormer in the primary cohort under within-subject CV with up to five folds and three seeds. Each dot averages one participant's available fold--seed scores; whiskers follow the $1.5\!\times\!\mathrm{IQR}$ rule. Sample sizes are $n\!=\!30,\,28,\,28,\,18$ at 20\%, 40\%, 60\% and 80\% MVC. Blue/red indicate male/female participants, respectively, corresponding to Supplementary Table~\ref{tab:sex_stratified}. Labels identify outliers by anonymized participant code.}
\label{fig:per_subject}
\end{figure}

\begin{table}[!htbp]
\centering
\caption{Within-subject outliers for full FatigueFormer in the primary cohort. A participant is flagged when their mean macro-F1 falls below $Q1 - 1.5\!\times\!\mathrm{IQR}$ of the corresponding MVC distribution in Supplementary Fig.~\ref{fig:per_subject}. Scores and cutoffs use the 0--1 scale; $n$ flagged counts outliers, not the full evaluation sample. P01--P30 are anonymized participant codes.}
\label{tab:per_subject_outliers}
\small
\setlength{\tabcolsep}{3pt}
\begin{tabular*}{\textwidth}{@{\extracolsep{\fill}}c p{0.58\textwidth} c c@{}}
\toprule
MVC (\%) & Outlier subject IDs (mean macro-F1) & $n$ flagged & Cutoff \\
\midrule
20 & P26 (0.314), P20 (0.813)                                       & 2 & 0.835 \\
40 & P12 (0.511), P20 (0.740), P07 (0.791), P08 (0.817), P09 (0.862) & 5 & 0.871 \\
60 & P18 (0.231), P08 (0.285)                                       & 2 & 0.446 \\
80 & P28 (0.439)                                                     & 1 & 0.716 \\
\bottomrule
\end{tabular*}
\end{table}

Supplementary Fig.~\ref{fig:per_subject} shows the participant-level spread behind the cohort means, with particularly broad variation at 60\% MVC. Supplementary Table~\ref{tab:per_subject_outliers} lists the low-scoring cases under the same distribution-based rule. These scores alone do not identify the causes of individual deviations.

Participants are not excluded because of low performance. The within-subject sample sizes reflect valid scores after constructing complete-context inputs, adapting fold counts to class support and skipping splits whose training sets lack a fatigue class (main-paper Section~\MainEvaluationSection{}). Flagged participants remain in the analyses wherever their inputs and splits are feasible (Supplementary Tables~\ref{tab:comparison} and~\ref{tab:generalization}).

\section*{S6. Exploratory sex-stratified performance}\label{sec:sex_stratified_si_section}

\begin{table}[!htbp]
\centering
\caption{Sex-stratified macro-F1 (\%) of full FatigueFormer in the primary cohort under within-subject and LOSO evaluation. M/F denote male/female groups. Values are bootstrap means with BCa 95\% CIs; $q$ values come from two-sided Mann--Whitney U tests with Benjamini--Hochberg correction across MVC levels within each protocol.}
\label{tab:sex_stratified}
\begin{tabular*}{\textwidth}{@{\extracolsep{\fill}}lllrlrr@{}}
\toprule
Panel & MVC (\%) & Group & $n$ & Mean F1 (95\% CI) & $q$ M vs.\ F (BH) & Effect $r$ \\
\midrule
\multirow{8}{*}{(a) within-subject} & \multirow{2}{*}{20} & M & 15 & 96.48 (94.35--97.42) & \multirow{2}{*}{0.499} & \multirow{2}{*}{$-0.33$} \\
                                    &                     & F & 15 & 89.46 (73.43--94.60) &                       &                          \\
\cmidrule(lr){2-7}
                                    & \multirow{2}{*}{40} & M & 13 & 93.91 (90.42--95.90) & \multirow{2}{*}{1.000} & \multirow{2}{*}{$-0.01$} \\
                                    &                     & F & 15 & 90.26 (80.20--94.87) &                       &                          \\
\cmidrule(lr){2-7}
                                    & \multirow{2}{*}{60} & M & 13 & 75.69 (58.43--86.80) & \multirow{2}{*}{1.000} & \multirow{2}{*}{$+0.13$} \\
                                    &                     & F & 15 & 87.56 (78.65--92.26) &                       &                          \\
\cmidrule(lr){2-7}
                                    & \multirow{2}{*}{80} & M &  9 & 91.09 (88.21--93.99) & \multirow{2}{*}{1.000} & \multirow{2}{*}{$+0.01$} \\
                                    &                     & F &  9 & 86.54 (69.10--93.34) &                       &                          \\
\midrule
\multirow{8}{*}{(b) LOSO}           & \multirow{2}{*}{20} & M & 15 & 61.36 (55.49--67.05) & \multirow{2}{*}{0.093} & \multirow{2}{*}{$-0.43$} \\
                                    &                     & F & 15 & 49.46 (40.12--59.11) &                       &                          \\
\cmidrule(lr){2-7}
                                    & \multirow{2}{*}{40} & M & 15 & 74.97 (63.16--83.21) & \multirow{2}{*}{0.320} & \multirow{2}{*}{$-0.22$} \\
                                    &                     & F & 15 & 69.71 (59.92--76.92) &                       &                          \\
\cmidrule(lr){2-7}
                                    & \multirow{2}{*}{60} & M & 15 & 52.63 (43.64--63.91) & \multirow{2}{*}{0.195} & \multirow{2}{*}{$+0.32$} \\
                                    &                     & F & 15 & 62.48 (53.11--70.54) &                       &                          \\
\cmidrule(lr){2-7}
                                    & \multirow{2}{*}{80} & M & 15 & 67.42 (58.66--74.40) & \multirow{2}{*}{0.072} & \multirow{2}{*}{$-0.51$} \\
                                    &                     & F & 15 & 52.87 (45.18--61.03) &                       &                          \\
\bottomrule
\end{tabular*}
\end{table}

Supplementary Table~\ref{tab:sex_stratified} reports subgroup estimates alongside the participant-level results in Section~S5. No sex contrast reaches the FDR-corrected threshold of $q<0.05$ under either protocol. With small subgroups and wide confidence intervals in several conditions, these exploratory comparisons do not establish equivalent performance between sexes.

\section*{S7. Component and feature analyses in SWIFTIES}\label{sec:swifties_si}

This section complements the analyses in main-paper Section~\MainResultsSection{} by comparing component effects and feature selection in SWIFTIES at 25\% and 45\% MVC with the primary cohort. These conditions provide an independent assessment in the low-to-moderate intensity range.

\paragraph*{S7.1 Component effects across protocols}
Supplementary Table~\ref{tab:sw_ablation} reports the SWIFTIES ablations, and Table~\ref{tab:component_matrix} summarizes significant effects across both cohorts. As in main-paper Figure~\MainAblationFigure{}, \emph{w/o cross-attn} replaces cross-attention with self-attention, while \emph{w/o multi-scale} omits the temporal and static auxiliary classification losses. The same definitions apply to the primary-cohort tables in Section~S8.

\begin{table}[!htbp]
\centering
\caption{SWIFTIES ablation results under within-subject and LOSO evaluation ($n\!=\!32$ at each MVC). Entries report mean macro-F1 (\%) followed by the BH-FDR-adjusted one-sided Wilcoxon $q$ value (or significance indicator) and matched-pairs rank-biserial $r$ versus the full model. $\star$: $q\!<\!0.05$ for full $>$ variant; n.s.: $q\geq0.05$; $\dagger$: the variant is significantly better than full in the reverse test. Negative $r$ favors the variant.}
\label{tab:sw_ablation}
\footnotesize
\setlength{\tabcolsep}{3pt}
\begin{tabular*}{\textwidth}{@{\extracolsep{\fill}}lcc@{}}
\toprule
\multicolumn{3}{c}{\textbf{(a) Within-subject five-fold cross-validation}} \\
\midrule
Variant & 25\% MVC & 45\% MVC \\
\midrule
FatigueFormer full & 87.57 & 83.68 \\
w/o static & 91.70 ($\dagger$; $-1.00$) & 87.92 ($\dagger$; $-0.92$) \\
w/o temporal & 74.61 ($9.3\!\times\!10^{-10}\,\star$; $1.00$) & 72.39 ($2.8\!\times\!10^{-9}\,\star$; $0.99$) \\
w/o cross-attn & 89.32 (n.s.; $-0.78$) & 85.77 (n.s.; $-0.74$) \\
w/o multi-scale & 84.94 ($3.3\!\times\!10^{-9}\,\star$; $0.98$) & 80.76 ($1.5\!\times\!10^{-6}\,\star$; $0.88$) \\
\midrule
\multicolumn{3}{c}{\textbf{(b) Leave-one-subject-out evaluation}} \\
\midrule
Variant & 25\% MVC & 45\% MVC \\
\midrule
FatigueFormer full & 58.02 & 66.62 \\
w/o static & 58.37 ($0.795$; $-0.08$) & 66.74 ($0.467$; $0.02$) \\
w/o temporal & 52.63 ($1.0\!\times\!10^{-5}\,\star$; $0.84$) & 59.54 ($1.6\!\times\!10^{-6}\,\star$; $0.89$) \\
w/o cross-attn & 58.90 ($0.795$; $-0.17$) & 65.32 ($0.374$; $0.17$) \\
w/o multi-scale & 57.87 ($0.795$; $0.03$) & 65.44 ($0.374$; $0.12$) \\
\bottomrule
\end{tabular*}
\end{table}

In SWIFTIES, removing the static branch improves within-subject performance at both intensities in the reverse tests, but produces no significant LOSO decrease in the forward tests. Removing the temporal branch significantly lowers performance under both protocols at both intensities. Replacing cross-attention with self-attention produces no significant decrease in the forward tests, while omitting auxiliary supervision significantly lowers performance only within subject. These results agree with the protocol-dependent roles described in the main paper: temporal context contributes across settings, whereas the other components do not benefit every condition. Statistical agreement between cohorts is incomplete: the temporal effect is significant at 45\% MVC LOSO in SWIFTIES, but the primary-cohort result at 40\% MVC has $q=0.062$. SWIFTIES does not test the primary cohort's high-intensity effects.

\begin{table}[!htbp]
\centering
\caption{Component effects across protocols, MVC levels and cohorts, derived from Tables~\ref{tab:sw_ablation}, \ref{tab:ablation_mvc} and~\ref{tab:ablation_loso}. $\star$: full exceeds the variant after BH-FDR correction ($q<0.05$); $\dagger$: the variant exceeds full in the reverse test; $\cdot$: no reported significant effect. The matrix summarizes within-condition tests, not tests of differences between intensities or cohorts.}
\label{tab:component_matrix}
\footnotesize
\setlength{\tabcolsep}{3pt}
\begin{tabular*}{\textwidth}{@{\extracolsep{\fill}}lcccccccccccc@{}}
\toprule
& \multicolumn{6}{c}{Within-subject CV} & \multicolumn{6}{c}{Leave-one-subject-out} \\
\cmidrule(lr){2-7}\cmidrule(lr){8-13}
& \multicolumn{4}{c}{Primary cohort} & \multicolumn{2}{c}{SWIFTIES} & \multicolumn{4}{c}{Primary cohort} & \multicolumn{2}{c}{SWIFTIES} \\
\cmidrule(lr){2-5}\cmidrule(lr){6-7}\cmidrule(lr){8-11}\cmidrule(lr){12-13}
Variant & 20\% & 40\% & 60\% & 80\% & 25\% & 45\% & 20\% & 40\% & 60\% & 80\% & 25\% & 45\% \\
\midrule
w/o static       & $\dagger$ & $\dagger$ & $\dagger$ & $\cdot$ & $\dagger$ & $\dagger$ & $\cdot$ & $\cdot$ & $\cdot$ & $\star$ & $\cdot$   & $\cdot$ \\
w/o temporal     & $\star$   & $\star$   & $\cdot$   & $\star$ & $\star$   & $\star$   & $\star$ & $\cdot$ & $\cdot$ & $\star$ & $\star$   & $\star$ \\
w/o cross-attn   & $\cdot$   & $\cdot$   & $\cdot$   & $\cdot$ & $\cdot$   & $\cdot$   & $\cdot$ & $\cdot$ & $\cdot$ & $\star$ & $\cdot$   & $\cdot$ \\
w/o multi-scale  & $\star$   & $\star$   & $\star$   & $\star$ & $\star$   & $\star$   & $\cdot$ & $\cdot$ & $\cdot$ & $\star$ & $\cdot$   & $\cdot$ \\
\bottomrule
\end{tabular*}
\end{table}

\paragraph*{S7.2 Feature-selection consistency and MI patterns across MVC}
The SWIFTIES stable core comprises 28 descriptors selected in every LOSO training fold at both intensities. It contains all 22 descriptors in the primary core plus six triceps descriptors: \texttt{TB\_AE}, \texttt{TB\_BE}, \texttt{TB\_BSE}, \texttt{TB\_SE}, \texttt{TB\_WENT} and \texttt{TB\_WA}. This is the subset relationship shown in main-paper Figure~3. The narrower intensity range in SWIFTIES may partly explain its larger core.

A separate comparison considers descriptors selected in at least 80\% of folds at each intensity. For the matched primary/SWIFTIES pairs of 20\%/25\% and 40\%/45\% MVC, Jaccard similarities between these sets are 0.857 and 0.875 across the 62-descriptor pool, and 1.000 and 0.889 when restricted to DEC descriptors. This comparison supports consistency of frequent selection at similar loads; it uses a less stringent criterion than the stable core.

To describe the amplitude- and power-related selection changes discussed in the main paper, we summarize MI for seven descriptors per muscle: \texttt{iEMG}, \texttt{RMS}, \texttt{MAV}, \texttt{MCV}, \texttt{DASDV}, \texttt{FSM2} and \texttt{TP}. At primary-cohort 20\%, 40\%, 60\% and 80\% MVC, biceps mean MI is 0.171370, 0.163231, 0.108640 and 0.045422; triceps mean MI is 0.029122, 0.046514, 0.040229 and 0.041996. Mean top-40 cutoffs are 0.033531, 0.215569, 0.251619 and 0.208025, respectively.

SWIFTIES means equally average the seven descriptors across 32 LOSO folds (224 MI values); cutoffs average each fold's 40th-highest MI. From 25\% to 45\% MVC, biceps MI decreases from 0.049767 to 0.042625 ($-14.35\%$), triceps MI from 0.036678 to 0.035538 ($-3.11\%$), and the cutoff rises from 0.043215 to 0.058190 ($+34.65\%$). Thus both cohorts show lower biceps MI and a higher selection cutoff from low to moderate intensity, while triceps trajectories differ. These summaries describe label association and competition within the fixed top-40 budget; they do not measure muscle activation or establish a recruitment mechanism. Individual selection frequencies, rather than a group's mean MI, determine whether its descriptors enter the stable core (Section~S9).

\clearpage
\section*{S8. Detailed primary-cohort performance and ablation tables}
Tables~\ref{tab:comparison} and~\ref{tab:generalization} provide the primary-cohort statistics behind main-paper Table~\MainPerformanceTable{}; Tables~\ref{tab:ablation_mvc} and~\ref{tab:ablation_loso} provide the ablation results behind Figure~\MainAblationFigure{}. The sample count $n$ is the number of participants with valid scores for that method and condition; paired tests use only participants shared by the two methods. Training and statistical procedures are detailed in Section~S3.
\begin{table}[!htbp]
\centering
\caption{Within-subject macro-F1 (\%) with BCa 95\% CIs where available. Full FatigueFormer uses training-fold MI top-40 selection and prototype loss; base retains all 62 descriptors and omits both additions. Boldface identifies the full reference model, not the best score. $q$: BH-FDR-adjusted one-sided paired Wilcoxon test for full $>$ comparator; $r$: matched-pairs rank-biserial effect size. RBF-SVM results are descriptive, without reported CIs or paired tests. Dashes denote unreported comparisons.}
\label{tab:comparison}
\begin{tabular*}{\textwidth}{@{\extracolsep{\fill}}llrlrr@{}}
\toprule
MVC (\%) & Method & $n$ & Mean F1 (95\% CI) & $q$ vs.\ full & Effect $r$ \\
\midrule
20 & \textbf{FatigueFormer full} & 30 & \textbf{92.97 (83.93--95.62)} & --                       & --   \\
   & FatigueFormer base          & 30 & 93.52 (84.37--96.06)          & --                       & --   \\
   & AutoInt                       & 30 & 86.31 (83.45--88.80)          & $2.4\!\times\!10^{-6}$   & 0.87 \\
   & RBF-SVM                       & 30 & 89.61                        & --                       & --   \\
   & CNN--SVM                      & 30 & 54.89 (49.83--60.42)          & $3.1\!\times\!10^{-9}$   & 1.00 \\
   & CLT                           & 30 & 33.24 (30.25--37.20)          & $3.1\!\times\!10^{-9}$   & 1.00 \\
   & MACNet                        & 30 & 24.62 (22.17--28.05)          & $3.1\!\times\!10^{-9}$   & 1.00 \\
\midrule
40 & \textbf{FatigueFormer full} & 28 & \textbf{91.95 (86.32--94.62)} & --                       & --   \\
   & FatigueFormer base          & 28 & 92.10 (86.31--94.82)          & --                       & --   \\
   & AutoInt                       & 29 & 85.56 (80.94--88.86)          & $1.1\!\times\!10^{-5}$   & 0.85 \\
   & RBF-SVM                       & 29 & 94.62                        & --                       & --   \\
   & CNN--SVM                      & 30 & 53.50 (47.21--59.71)          & $1.2\!\times\!10^{-8}$   & 1.00 \\
   & CLT                           & 30 & 24.23 (21.70--29.45)          & $9.3\!\times\!10^{-9}$   & 1.00 \\
   & MACNet                        & 30 & 22.13 (19.90--28.19)          & $9.3\!\times\!10^{-9}$   & 1.00 \\
\midrule
60 & \textbf{FatigueFormer full} & 28 & \textbf{82.05 (72.27--87.95)} & --                       & --   \\
   & FatigueFormer base          & 28 & 81.79 (72.18--87.69)          & --                       & --   \\
   & AutoInt                       & 30 & 78.62 (74.09--82.27)          & $0.015$                  & 0.48 \\
   & RBF-SVM                       & 30 & 92.59                        & --                       & --   \\
   & CNN--SVM                      & 30 & 43.47 (38.53--49.25)          & $3.1\!\times\!10^{-8}$   & 0.99 \\
   & CLT                           & 30 & 18.61 (17.66--20.94)          & $1.9\!\times\!10^{-8}$   & 1.00 \\
   & MACNet                        & 30 & 20.03 (18.78--21.51)          & $1.9\!\times\!10^{-8}$   & 1.00 \\
\midrule
80 & \textbf{FatigueFormer full} & 18 & \textbf{88.81 (78.72--92.35)} & --                       & --   \\
   & FatigueFormer base          & 18 & 87.16 (78.69--91.22)          & --                       & --   \\
   & AutoInt                       & 26 & 71.49 (64.92--77.71)          & $9.9\!\times\!10^{-4}$   & 0.80 \\
   & RBF-SVM                       & 26 & 92.75                        & --                       & --   \\
   & CNN--SVM                      & 26 & 31.04 (27.71--36.68)          & $6.4\!\times\!10^{-6}$   & 1.00 \\
   & CLT                           & 26 & 17.13 (16.46--17.73)          & $6.4\!\times\!10^{-6}$   & 1.00 \\
   & MACNet                        & 26 & 19.28 (18.12--20.58)          & $6.4\!\times\!10^{-6}$   & 1.00 \\
\bottomrule
\end{tabular*}
\end{table}

\begin{table}[!htbp]
\centering
\caption{LOSO macro-F1 (\%) with BCa 95\% CIs across 30 held-out participants at each MVC. Neural models use 28 training participants and one validation participant per fold; RBF-SVM uses all 29 non-test participants without a validation holdout. Boldface identifies full FatigueFormer. $q$ and $r$ follow the paired-test definitions in Table~\ref{tab:comparison}; here RBF-SVM CIs and paired tests are available.}
\label{tab:generalization}
\begin{tabular*}{\textwidth}{@{\extracolsep{\fill}}llrlrr@{}}
\toprule
MVC (\%) & Method & $n$ & Mean F1 (95\% CI) & $q$ vs.\ full & Effect $r$ \\
\midrule
20 & \textbf{FatigueFormer full} & 30 & \textbf{55.41 (49.06--61.18)} & --                       & --   \\
   & FatigueFormer base          & 30 & 53.59 (48.14--58.71)          & --                       & --   \\
   & AutoInt                       & 30 & 41.21 (35.60--47.08)          & $1.2\!\times\!10^{-4}$   & 0.73 \\
   & RBF-SVM                           & 30 & 42.01 (36.77--47.50)          & $1.2\!\times\!10^{-4}$   & 0.74 \\
   & CNN--SVM                      & 30 & 34.60 (30.19--39.21)          & $1.7\!\times\!10^{-7}$   & 0.94 \\
   & CLT                           & 30 & 24.15 (20.23--28.31)          & $4.4\!\times\!10^{-8}$   & 0.97 \\
   & MACNet                        & 30 & 17.81 (14.38--21.77)          & $9.3\!\times\!10^{-9}$   & 1.00 \\
\midrule
40 & \textbf{FatigueFormer full} & 30 & \textbf{72.34 (65.22--78.19)} & --                       & --   \\
   & FatigueFormer base          & 30 & 72.52 (66.30--77.91)          & --                       & --   \\
   & AutoInt                       & 30 & 51.63 (43.83--59.36)          & $3.0\!\times\!10^{-6}$   & 0.86 \\
   & RBF-SVM                           & 30 & 51.47 (44.26--57.47)          & $2.4\!\times\!10^{-6}$   & 0.87 \\
   & CNN--SVM                      & 30 & 37.57 (32.23--42.84)          & $1.6\!\times\!10^{-8}$   & 0.98 \\
   & CLT                           & 30 & 26.52 (20.07--33.67)          & $4.7\!\times\!10^{-9}$   & 1.00 \\
   & MACNet                        & 30 & 10.88 (8.23--16.19)           & $4.7\!\times\!10^{-9}$   & 1.00 \\
\midrule
60 & \textbf{FatigueFormer full} & 30 & \textbf{57.56 (50.92--64.47)} & --                       & --   \\
   & FatigueFormer base          & 30 & 59.55 (53.15--65.27)          & --                       & --   \\
   & AutoInt                       & 30 & 43.79 (38.26--50.22)          & $3.9\!\times\!10^{-3}$   & 0.56 \\
   & RBF-SVM                           & 30 & 46.13 (39.62--53.26)          & $0.014$                  & 0.46 \\
   & CNN--SVM                      & 30 & 32.92 (27.68--38.57)          & $6.7\!\times\!10^{-6}$   & 0.85 \\
   & CLT                           & 30 & 26.81 (20.92--34.38)          & $5.5\!\times\!10^{-6}$   & 0.87 \\
   & MACNet                        & 30 & 21.82 (16.69--29.46)          & $4.1\!\times\!10^{-7}$   & 0.94 \\
\midrule
80 & \textbf{FatigueFormer full} & 30 & \textbf{60.15 (53.92--66.32)} & --                       & --   \\
   & FatigueFormer base          & 30 & 55.24 (48.49--62.44)          & --                       & --   \\
   & AutoInt                       & 30 & 36.49 (30.71--42.94)          & $5.8\!\times\!10^{-6}$   & 0.85 \\
   & RBF-SVM                           & 30 & 40.52 (34.25--48.03)          & $1.7\!\times\!10^{-4}$   & 0.71 \\
   & CNN--SVM                      & 30 & 31.04 (25.36--36.18)          & $3.2\!\times\!10^{-7}$   & 0.94 \\
   & CLT                           & 30 & 23.80 (18.82--30.06)          & $3.2\!\times\!10^{-7}$   & 0.94 \\
   & MACNet                        & 30 & 25.68 (18.86--34.23)          & $2.7\!\times\!10^{-6}$   & 0.88 \\
\bottomrule
\end{tabular*}
\end{table}

\clearpage
\begin{table}[H]
\centering
\caption{Within-subject ablations: mean macro-F1 (\%) with BCa 95\% CIs. Tests compare full $>$ variant using BH-FDR-adjusted one-sided paired Wilcoxon tests; $r$ is the matched-pairs rank-biserial effect size. $\star$: $q<0.05$; n.s.: $q\geq0.05$ in the forward test; $\dagger$: a significant reverse test favors the variant. Thus n.s.\ and $\dagger$ can appear together. Negative $r$ favors the variant. Ablation definitions are given in Section~S7.1.}
\label{tab:ablation_mvc}
\begin{tabular*}{\textwidth}{@{\extracolsep{\fill}}llrlrr@{}}
\toprule
MVC (\%) & Variant & $n$ & Mean F1 (95\% CI) & $q$ vs.\ full & Effect $r$ \\
\midrule
20 & FatigueFormer full & 30 & 92.97 (83.93--95.62) & --                                & --      \\
   & w/o static                & 30 & 94.19 (85.20--96.68) & n.s.~$\dagger$                    & $-0.85$ \\
   & w/o temporal              & 30 & 85.55 (78.80--89.04) & $4.7\!\times\!10^{-9}\,\star$     & 1.00    \\
   & w/o cross-attn            & 30 & 93.72 (84.65--96.27) & n.s.                              & $-0.90$ \\
   & w/o multi-scale           & 30 & 91.65 (83.03--94.40) & $3.3\!\times\!10^{-8}\,\star$     & 0.97    \\
\midrule
40 & FatigueFormer full & 28 & 91.95 (86.32--94.62) & --                                & --      \\
   & w/o static                & 28 & 93.34 (88.44--95.58) & n.s.~$\dagger$                    & $-0.61$ \\
   & w/o temporal              & 28 & 88.43 (82.78--92.06) & $1.2\!\times\!10^{-4}\,\star$     & 0.81    \\
   & w/o cross-attn            & 28 & 92.63 (86.97--94.97) & n.s.                              & $-0.55$ \\
   & w/o multi-scale           & 28 & 90.87 (84.72--93.85) & $1.5\!\times\!10^{-4}\,\star$     & 0.78    \\
\midrule
60 & FatigueFormer full & 28 & 82.05 (72.27--87.95) & --                                & --      \\
   & w/o static                & 28 & 83.30 (73.57--88.95) & n.s.~$\dagger$                    & $-0.48$ \\
   & w/o temporal              & 28 & 81.53 (72.51--87.07) & n.s.                              & 0.35    \\
   & w/o cross-attn            & 28 & 81.57 (71.61--87.59) & n.s.                              & 0.20    \\
   & w/o multi-scale           & 28 & 81.34 (71.35--87.12) & $0.039\,\star$                    & 0.52    \\
\midrule
80 & FatigueFormer full & 18 & 88.81 (78.72--92.35) & --                                & --      \\
   & w/o static                & 18 & 88.53 (79.16--91.94) & n.s.                              & 0.29    \\
   & w/o temporal              & 18 & 86.71 (79.94--90.74) & $0.043\,\star$                    & 0.57    \\
   & w/o cross-attn            & 18 & 88.60 (78.38--92.20) & n.s.                              & 0.32    \\
   & w/o multi-scale           & 18 & 87.06 (77.20--90.96) & $0.043\,\star$                    & 0.58    \\
\bottomrule
\end{tabular*}
\end{table}
The reverse-direction tests for static-branch removal at 20\%, 40\% and 60\% MVC give $q<10^{-4}$, $q=0.001$ and $q=0.013$, respectively. These correspond to the $\dagger$ entries above; the signed effect sizes remain defined in the full-minus-variant direction.

\begin{table}[H]
\centering
\caption{LOSO ablations across 30 held-out participants at each MVC. Values are mean macro-F1 (\%) with BCa 95\% CIs. Statistical conventions follow Table~\ref{tab:ablation_mvc}: $\star$ marks a significant decrease from full ($q<0.05$), and negative $r$ favors the variant in the paired rank comparison.}
\label{tab:ablation_loso}
\begin{tabular*}{\textwidth}{@{\extracolsep{\fill}}llrlrr@{}}
\toprule
MVC (\%) & Variant & $n$ & Mean F1 (95\% CI) & $q$ vs.\ full & Effect $r$ \\
\midrule
20 & FatigueFormer full & 30 & 55.41 (49.06--61.18) & --                                & --     \\
   & w/o static                 & 30 & 53.11 (47.15--58.48) & 0.142                             & 0.29   \\
   & w/o temporal               & 30 & 50.91 (45.82--55.81) & $0.013\,\star$                    & 0.57   \\
   & w/o cross-attn             & 30 & 52.93 (46.90--58.27) & 0.100                             & 0.37   \\
   & w/o multi-scale            & 30 & 54.37 (48.66--59.47) & 0.205                             & 0.21   \\
\midrule
40 & FatigueFormer full & 30 & 72.34 (65.22--78.19) & --                                & --     \\
   & w/o static                 & 30 & 73.16 (66.72--78.84) & n.s.                              & $-0.19$\\
   & w/o temporal               & 30 & 68.93 (61.76--74.84) & 0.062                             & 0.47   \\
   & w/o cross-attn             & 30 & 73.24 (65.41--79.73) & n.s.                              & $-0.11$\\
   & w/o multi-scale            & 30 & 73.23 (65.65--79.16) & n.s.                              & $-0.19$\\
\midrule
60 & FatigueFormer full & 30 & 57.56 (50.92--64.47) & --                                & --     \\
   & w/o static                 & 30 & 58.07 (51.97--64.29) & n.s.                              & $-0.05$\\
   & w/o temporal               & 30 & 57.37 (50.96--63.40) & n.s.                              & 0.05   \\
   & w/o cross-attn             & 30 & 59.73 (52.12--66.98) & n.s.                              & $-0.20$\\
   & w/o multi-scale            & 30 & 58.08 (51.13--64.72) & n.s.                              & 0.01   \\
\midrule
80 & FatigueFormer full & 30 & 60.15 (53.92--66.32) & --                                & --     \\
   & w/o static                 & 30 & 52.72 (45.92--59.75) & $0.0048\,\star$                   & 0.57   \\
   & w/o temporal               & 30 & 52.35 (44.56--59.93) & $0.011\,\star$                    & 0.49   \\
   & w/o cross-attn             & 30 & 50.89 (43.12--58.89) & $0.0024\,\star$                   & 0.66   \\
   & w/o multi-scale            & 30 & 51.82 (44.24--60.21) & $0.0048\,\star$                   & 0.59   \\
\bottomrule
\end{tabular*}
\end{table}

\clearpage
\section*{S9. Attention visualizations and feature-selection frequencies}
These primary-cohort maps complement the analyses in main-paper Section~\MainResultsSection{}.

\begin{figure}[H]\centering\includegraphics[width=\textwidth]{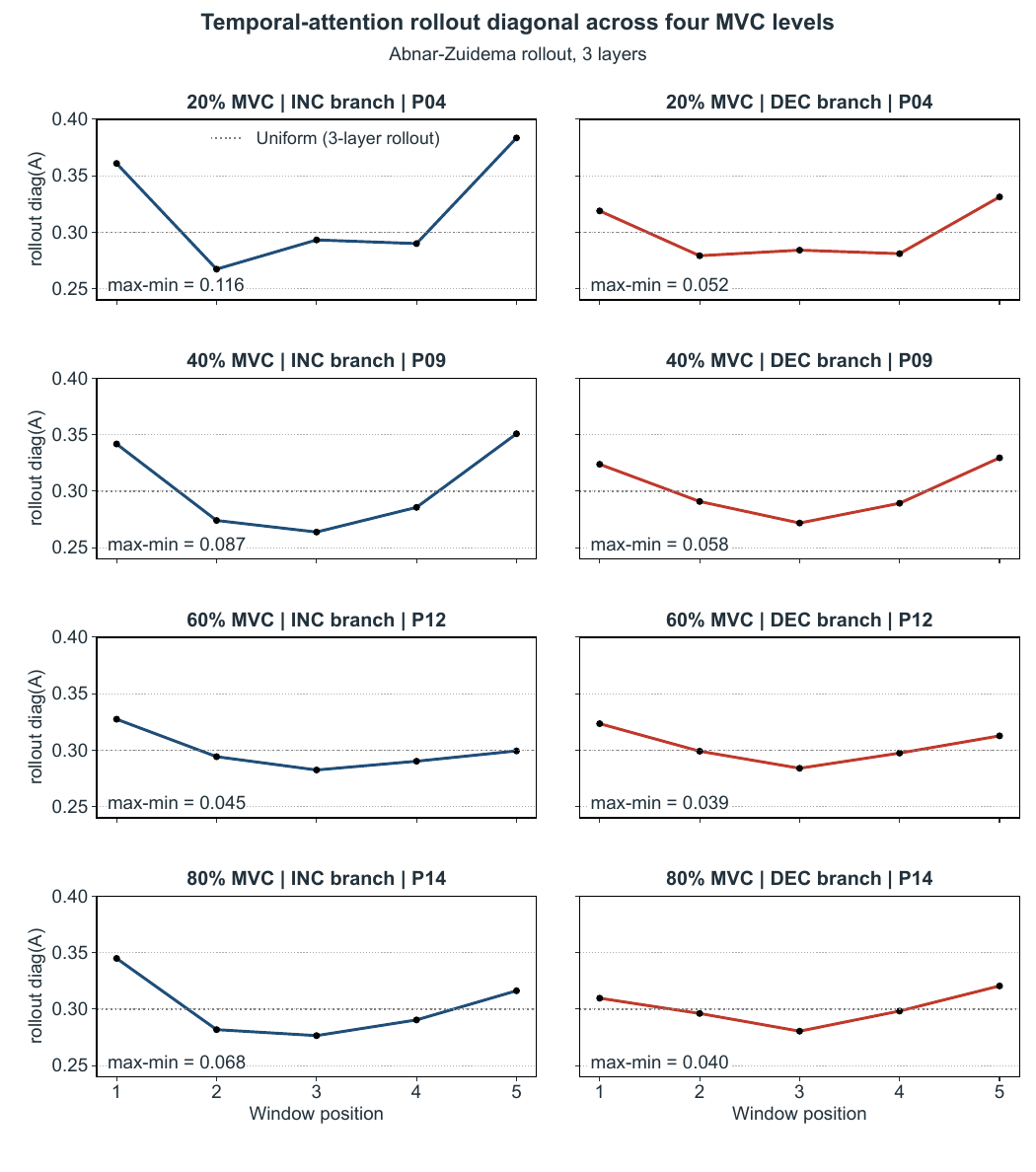}\caption{Temporal rollout diagonals for full FatigueFormer's median-performing LOSO fold at each MVC. Positions 1--5 index consecutive windows; position 3 is the labeled target. Panels identify the INC/DEC stream and held-out participant. The dotted line is the uniform three-layer rollout reference; max--min summarizes variation across positions.}\label{fig:temporal_attention}\end{figure}

\begin{figure}[H]\centering\includegraphics[width=\textwidth]{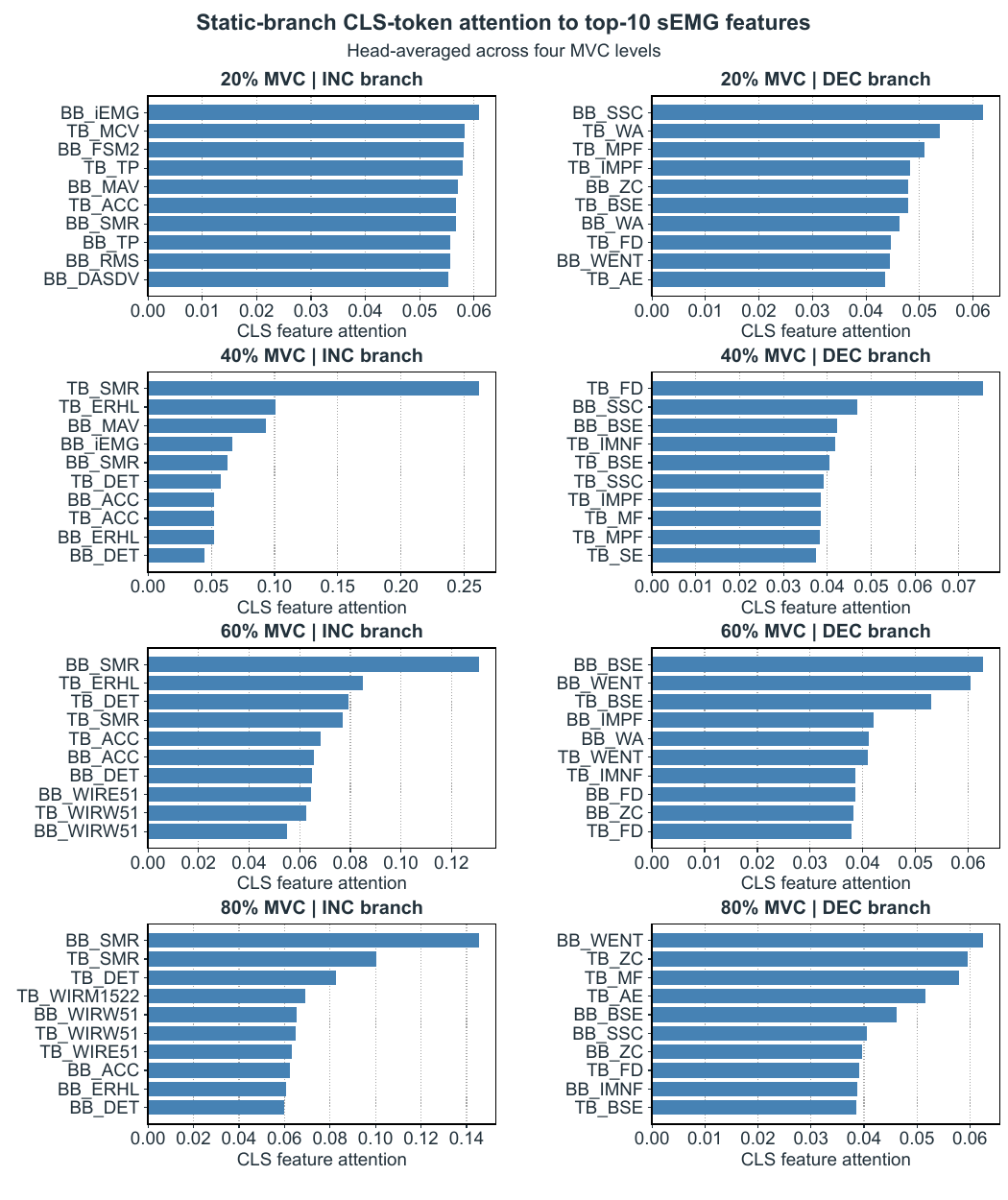}\caption{Head-averaged static CLS-token attention to the ten highest-weighted descriptors per stream in the median-performing LOSO folds. Rankings describe attention within the MI-selected pool, not causal feature importance. BB/TB denote biceps/triceps; definitions are in Table~\ref{tab:feature_definitions}. Horizontal scales vary between panels. Component contributions are assessed by ablation.}\label{fig:static_attention}\end{figure}

\begin{figure}[H]\centering\includegraphics[width=0.97\textwidth]{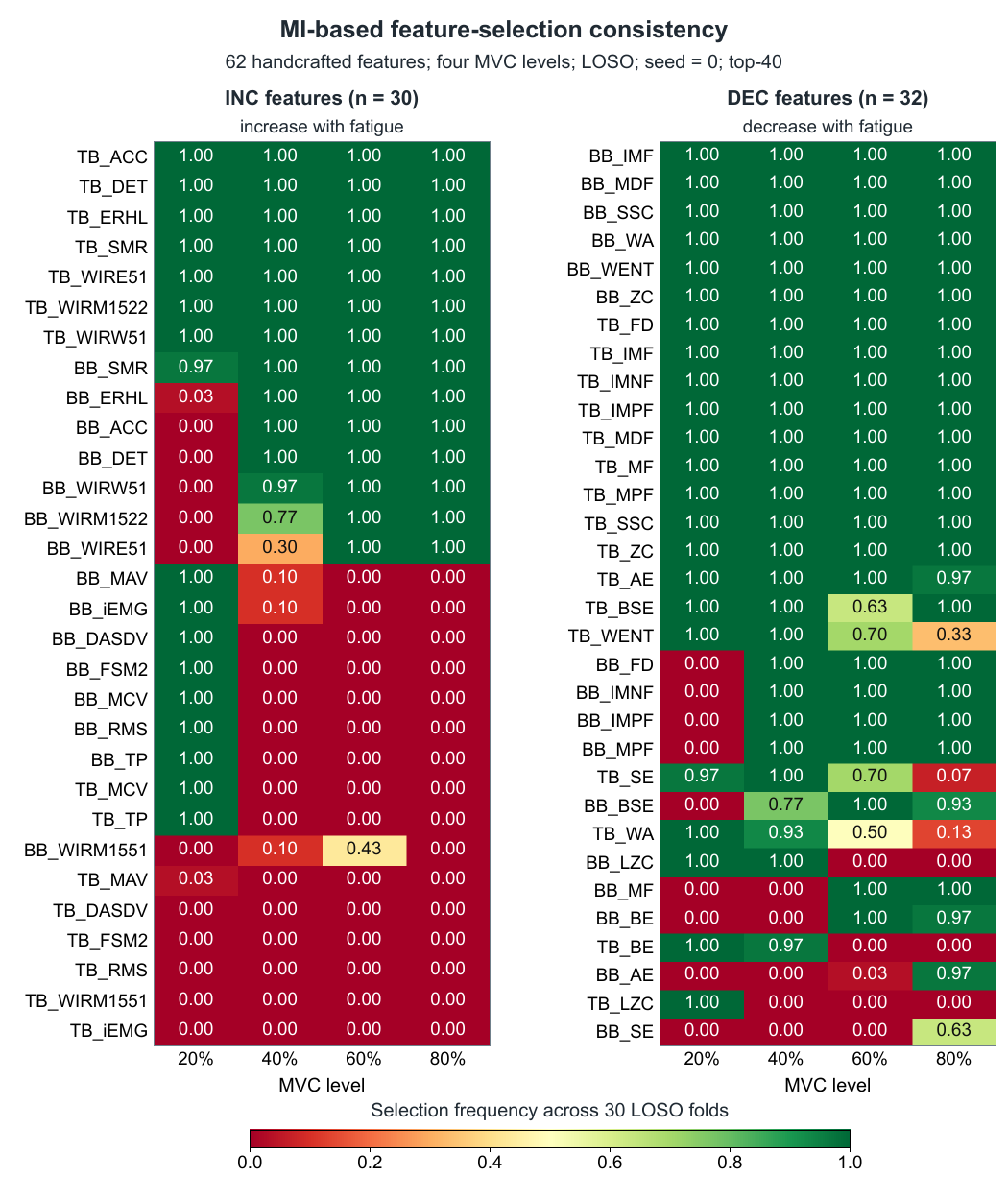}\caption{MI top-40 selection frequencies for all 62 muscle-specific descriptors across 30 primary-cohort LOSO training folds per MVC. Each cell is the fraction of folds selecting that descriptor, displayed to two decimal places. The 22 rows with frequency 1.00 at all four intensities form the primary stable core. INC/DEC panel counts refer to descriptors, not participants.}\label{fig:selection_frequency}\end{figure}

\bibliographystyle{ieee}
\bibliography{references}